\documentclass[letterpaper]{article} 
\usepackage{aaai25}  
\usepackage{times}  
\usepackage{helvet}  
\usepackage{courier}  
\usepackage[hyphens]{url}  
\usepackage{graphicx} 
\usepackage{natbib}  
\usepackage{caption} 
\usepackage{amsmath}
\usepackage{hyperref}
\usepackage{cleveref}
\usepackage{graphicx}
\usepackage{booktabs}
\usepackage{bm}
\usepackage{pifont}       
\newcommand{\cmark}{\ding{51}}
\newcommand{\xmark}{\ding{55}}

\usepackage{array}
\usepackage{ragged2e} 
\usepackage{tabularx}
\usepackage{colortbl}
\usepackage{multirow}
\usepackage{multicol}
\definecolor{mygray}{gray}{.9}
\usepackage{amssymb}
\usepackage[linesnumbered,ruled,vlined]{algorithm2e}
\SetKwInput{KwInit}{Initialization}
\SetKwInput{Kw}{}

\nocopyright

\title{Target Semantics Clustering via Text Representations

for Robust Universal Domain Adaptation}
\author{
    Weinan He\textsuperscript{\rm 1}, Zilei Wang\textsuperscript{\rm 1}\thanks{Corresponding author.}, Yixin Zhang\textsuperscript{\rm 1,2}\\
}
\affiliations{
    \textsuperscript{\rm 1}University of Science and Technology of China, Hefei, China\\

    \textsuperscript{\rm 2}Institute of Artificial Intelligence, Hefei Comprehensive National Science Center\\
    hwn2018@mail.ustc.edu.cn, \{zlwang, zhyx12\}@ustc.edu.cn
}

\newcommand{\ie}{i.e.}

\begin{document}

\maketitle

\begin{abstract}
Universal Domain Adaptation (UniDA) focuses on transferring source domain knowledge to the target domain under both domain shift and unknown category shift. Its main challenge lies in identifying common class samples and aligning them.
Current methods typically obtain target domain semantics centers from an unconstrained continuous image representation space. Due to domain shift and the unknown number of clusters, these centers often result in complex and less robust alignment algorithm.
In this paper, based on vision-language models, we search for semantic centers in a semantically meaningful and discrete text representation space. The constrained space ensures almost no domain bias and appropriate semantic granularity for these centers, enabling a simple and robust adaptation algorithm.
Specifically, we propose TArget Semantics Clustering (TASC) via Text Representations, which leverages information maximization as a unified objective and involves two stages.
First, with the frozen encoders, a greedy search-based framework is used to search for an optimal set of text embeddings to represent target semantics.
Second, with the search results fixed, encoders are refined based on gradient descent, simultaneously achieving robust domain alignment and private class clustering.
Additionally, we propose Universal Maximum Similarity (UniMS), a scoring function tailored for detecting open-set samples in UniDA. 
Experimentally, we evaluate the universality of UniDA algorithms under four category shift scenarios. Extensive experiments on four benchmarks demonstrate the effectiveness and robustness of our method, which has achieved state-of-the-art performance.
\end{abstract}
\begin{links}
    \link{Code}{https://github.com/Sapphire-356/TASC}
\end{links}

\section{Introduction}
\label{sec:intro}

\begin{figure}[!t]
	\centering
	\includegraphics[scale=0.6]{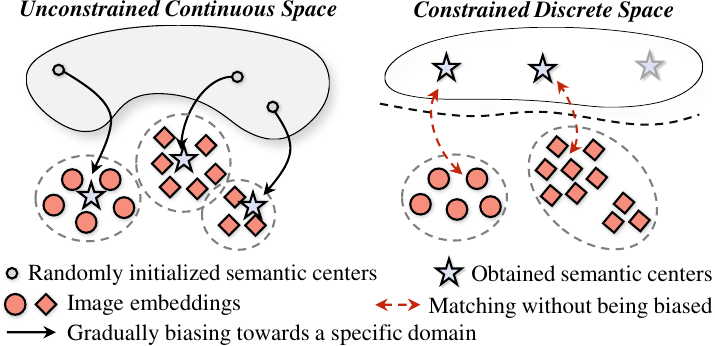}
	\caption{Illustration of our core idea. The left part abstractly represents the updating process of semantic centers in existing clustering-based UniDA methods. These centers gradually become domain-specific and have uncontrollable semantic granularity, such as separating diamonds into two categories. In the right part, we constrain the search space of semantic centers to a semantically meaningful and discrete space, alleviating the issues of domain bias and inappropriate semantic granularity.}
	\label{fig:Intro}
\end{figure}

Deep neural networks have achieved remarkable success across various computer vision tasks~(\citeauthor{carion2020end,dosovitskiy2020image,he2016deep,imagenet,he2017mask}). 
However, the high cost of annotated data and the limitation of the independent and identically distributed (i.i.d.) assumptions between training and test datasets pose challenges in practical applications. 
To tackle these issues, Unsupervised Domain Adaptation~(DA)~(\citeauthor{pan2009survey,ben2010theory,ganin2015unsupervised,long2015learning,uda-pcl,zhang2023towards,li2021semantic,zhang2022low,gao2023exploit}) has emerged, aiming to transfer models trained on labeled source domains to unlabeled target domains.
Although DA has shown success, traditional DA methods rely on the closed-set assumption, which assumes that the source and target domains share the same label set, and this assumption can be easily violated in real-world scenarios. 
In light of this, Open-set DA~(ODA)~\cite{saito2018open} and Partial DA~(PDA)~\cite{cao2018partial} consider the existence of private classes in the target and source domains, respectively.
However, in ODA and PDA, prior knowledge about the locations of private classes is still required. 
To adapt to more general scenarios, Universal Domain Adaptation~(UniDA)~\cite{unida_UAN} has been proposed, aiming to achieve DA without any prior knowledge of category shifts, \ie, UniDA methods should be able to handle Closed-set DA (CDA), ODA, PDA, and Open-Partial DA (OPDA) simultaneously.

Currently, researchers have proposed many solutions, among which the target domain clustering based approach has been widely adopted~\cite{unida_DANCE,unida_DCC,unida_UniOT,unida_GLC}.
Target domain clustering can effectively mine the intrinsic structure of the target domain to enhance discriminability, which is almost immune to category shift.
Nevertheless, a more critical issue in UniDA is common class detection for cross-domain alignment.
In existing methods, the prototype-based classifier in the source domain and the cluster centers in the target domain, both obtained from a continuous image representation space, are expected to accurately represent the corresponding semantic categories.
Based on these semantic centers, common class detection is achieved through cluster-level matching~\cite{unida_DCC} or sample-semantic center matching~\cite{unida_DANCE,unida_UniOT,unida_GLC}, which are all based on representation similarity.
However, these semantic centers are actually difficult to use for UniDA: 1) they are domain biased; 2) the number of target domain clusters (or semantic granularity) is hard to estimate.
The former makes the representation similarity unreliable and challenging to discriminate common classes across domains. This leads to complex matching and alignment mechanisms~\cite{unida_UniOT,unida_DCC}.
The latter causes the algorithm less robust when facing different category shift scenarios as we cannot determine which domain has private classes and how many there are. Consequently,~\cite{unida_UniOT,unida_GLC} are compelled to assume that private classes must exist in the target domain.

Recently, the emergence of Vision-Language Models~(VLMs), such as CLIP~\cite{CLIP}, has offered a promising alternative for visual representation learning.
Benefiting from training on web-scale image-text pairs, CLIP exhibits strong cross-modal matching capabilities.
The encoded text representations of CLIP possess a certain extent of domain generalization.
Moreover, it provides decent representations for open-world images. 
Therefore, we believe that CLIP provides significant advantages for addressing domain shift and detecting private samples in UniDA.
However, how to better leverage these capabilities in CLIP to address the UniDA problem remains largely unexplored.

Considering the above analysis, in this paper, our core idea (Fig.~\ref{fig:Intro}) is that uniformly representing the image semantics of both domains in the text representation space to facilitate simple and robust UniDA algorithms on CLIP. 
For source domain, we obtain a set of embeddings from class names, as its semantic centers. 
For target domain, we search for a set of optimal text embeddings from the semantically meaningful and discrete text representation space. 
Within the constrained search space, our algorithms ensure that the two sets of embeddings possess the following properties: 1) almost no domain bias, and the embeddings between common classes across two domains are identical or very close; 2) appropriate semantic granularity. 
These properties enable: 1) simple matching of common classes through similarity and unifying the losses of common class alignment and private class clustering; 2) simple estimation of the number of private classes, and being aware of category shift.

Specifically, we propose TArget Semantics Clustering (TASC) via Text Representations.
It employs information maximization as a unified optimization objective to robustly adapt the model to the target domain in the presence of unknown category shifts. 
Mathematically, TASC is formulated as a Mixed-Integer Nonlinear Programming problem, which we solve through a two-stage optimization process.
In the first stage, the continuous parameters of encoders are fixed, and a set of text embeddings representing the target semantics is searched from a semantically meaningful and discrete text representation space based on a greedy search framework. 
In the second stage, the discrete variables in the search results are fixed, and the encoders in the continuous parameter space are further optimized based on gradient descent, achieving robust domain adaptation and target private class clustering simultaneously. 
Additionally, based on the semantic centers of both domains, we explicitly model the category shift. Benefiting from this, we design the Universal Maximum Similarity (UniMS), which is capable of perceiving category shift and is tailored for detecting target private sample in the UniDA task. We optimize the Gaussian Mixture Model to obtain the optimal threshold instead of using hand-tuned ones as in previous approaches.

Experimentally, we simultaneously evaluate UniDA algorithms under three types of category shift scenarios (OPDA, ODA, PDA) and one non-category shift scenario (CDA). We compare our method with existing ones using H-score metric on four common benchmarks. Extensive experimental results demonstrate that our method is sufficiently robust and achieves state-of-the-art performance.

Our contributions are summarized as follows:
\begin{itemize}
	\item We propose to uniformly represent the image semantics in the semantically rich and discrete text representation space to facilitate simple and robust UniDA algorithms.
    \item We propose TArget Semantics Clustering (TASC) via Text Representations, which employs information maximization as a unified objective and involves two optimization stages. It achieves robust domain adaptation and target private class clustering.
    \item We propose Universal Maximum Similarity (UniMS), which is capable of perceiving category shifts and accurately detecting target private samples.
    \item We evaluate UniDA algorithms under four different category shift scenarios. Our method is robust and achieves state-of-the-art performance on four benchmarks.
\end{itemize}

\section{Related Work}

\subsubsection{Universal Domain Adaptation.}
UniDA~\cite{unida_UAN} aims to address the domain adaptation without prior knowledge of label set relationship. 
\cite{unida_UAN,unida_CMU,unida_UniAM} propose multiple criteria for unknown detection.
\cite{unida_OVANet,unida_OneRing} design the special classifiers.
In~(\citeauthor{unida_GATE,unida_MATHS,unida_DANCE,unida_TNT,unida_MLNet}), neighborhood structures are exploited.
\cite{unida_uniood} explores the foundation models for UniDA.
\cite{unida_COCA} address UniDA with few-shot settings.
Recently, target domain clustering~\cite{unida_GLC,unida_UniOT,unida_DCC} has been developed to discover target domain categories and detect common class samples.
However, all these methods search for semantic centers in the sub-optimal unconstrained continuous image representation space.
In contrast, we explore the semantically meaningful and discrete text representation space.

\subsubsection{Vision Task with Vision-language Models.}
Recently, Vision-language Models (VLMs) have attracted increasing attention in multiple vision tasks~(\citeauthor{zhang2024vision,CLIP,yao2021filip,li2022grounded}). In this paper, we focus on the image clustering~\cite{cai2023semantic,li2023image,joseph2022novel} and open-set adaptation~\cite{min2023uota,zara2023autolabel,yu2023open}.
\cite{min2023uota} leveraging open-set unlabeled data in the wild for open-set task adaptation.
\cite{zara2023autolabel} consider the adaptation of action recognition model in the open-set scenario.
\cite{yu2023open} explore the potential of CLIP for ODA.
\cite{cai2023semantic,li2023image} propose leveraging external knowledge from text modality to facilitate clustering.
Moreover,~\cite{SCD} introduces a new task of obtaining class names for unlabeled datasets.
In this paper, we emphasize the role of the semantically meaningful text representation space in developing a simple and robust UniDA method.

\section{Preliminary}

Before we describe the details of our method, we firstly present the preliminaries used in our framework and formalize Universal Domain Adaptation (UniDA).

\noindent\textbf{Model.} 
In this paper, we focus on adapting the vision-language model CLIP~\cite{CLIP} to target domain in the UniDA settings. CLIP consists of image encoder $f$ and text encoder $g$. 
For a given image $\mathbf{x}$ and a set of class names $\mathcal{T}^s=\{t_1^s, t_2^s, \dots, t_m^s\} $, CLIP can make prediction by comparing image embedding with the text embeddings. 
Let's denote $\mathbf{z}=f(\mathbf{x})$ and $\mathbf{s}_j=g(t_j^s)$ as the $L_2$-normalized embeddings of the image and class name $j$ respectively. Then, the probability that $\mathbf{x}$ belongs to class $i$ is calculated as
\begin{equation}
\label{eqn:clipzeroshot}
    p_i = P(\mathbf{s}_i|\mathbf{z};\tau) = \frac{\text{exp}(\text{sim}(\mathbf{z}, \mathbf{s}_i) / \tau)}{\sum^{m}_{j=1}\text{exp}(\text{sim}(\mathbf{z}, \mathbf{s}_j) / \tau)},
\end{equation}
where $\text{sim}(\cdot, \cdot)$ is the cosine similarity and $\tau$ is the softmax temperature. For simplicity, let define the function $h$ as:
\begin{equation}
\label{eqn:func_h}
    h(\mathbf{z};\mathbf{S},\tau) \triangleq [p_1,p_2,\dots,p_m]^T = \mathbf{p} ,
\end{equation}
where $\mathbf{S}=[\mathbf{s}_1, \mathbf{s}_2, \dots, \mathbf{s}_m]$.
Also for simplicity, we omit the prompting strategy as default. Actually, we use the ensemble text templates from~(\citeauthor{lin2023multimodality}).
To efficiently transfer the CLIP model to the target domain, inspired by~(\citeauthor{smith2023construct, doveh2023teaching, cascante2023going}), we fine-tune both the image encoder $f$ and text encoder $g$ via LoRA~(\citeauthor{hu2021lora}).

\noindent\textbf{Universal Domain Adaptation.}
In the UniDA problem, we are given a labeled source domain $\mathcal{D}^s=\{(\mathbf{x}_i^s,y_i^s)\}_{i=1}^{n_s}$ and an unlabeled target domain $\mathcal{D}^t=\{\mathbf{x}_i^t\}_{i=1}^{n_t}$, where a domain gap exists between them.
We denote $\mathcal{C}_s$ and $\mathcal{C}_t$ as the label sets of source and target domain respectively.
Additionally, let $\mathcal{T}^s$ and $\mathcal{T}^t$ denote the sets of the class names.
Notably, we lack any prior information about $\mathcal{C}_t$ and $\mathcal{T}^t$ during training.
The common label set is represented by $\mathcal{C}=\mathcal{C}_s \,\cap\, \mathcal{C}_t$. 
Let $\overline{\mathcal{C}}_s=\mathcal{C}_s\backslash\mathcal{C}$ and $\overline{\mathcal{C}}_t=\mathcal{C}_t\backslash\mathcal{C}$ as label sets of source-private and target-private, respectively.
UniDA aims to train a model on $\mathcal{D}^s$ and $\mathcal{D}^t$ that can accurately classifies the target domain common class samples into $\lvert\mathcal{C}\rvert$ classes and, if target-private class samples exist, assigns them to a single \textit{unknown} class.

\noindent\textbf{Notations.}
First, all vectors in this paper are column vectors. 
When a matrix is constructed from vectors, these vectors are organized in a column-wise manner. 
Let $\mathcal{\mathbf{W}}^s=[\mathbf{w}_1^s, \mathbf{w}_2^s, \dots, \mathbf{w}_{\lvert\mathcal{C}_s\rvert}^s]$ denote the text embeddings of $\mathcal{T}^s$, where $\mathbf{w}^s_j = g(t_j^s)$.
Additionally, Given the prototypes and temperature, the entropy of an embedding can be defined based on predicted probabilities using prototypes. For simplicity, on the basis of \mbox{Eq. (\ref{eqn:func_h})}, we define the this function:
\begin{equation}
\label{eqn:func_ent}
    \text{Entropy}(\mathbf{p}) \triangleq - \sum_{i=1}^m p_i \log p_i,
\end{equation}
where $\mathbf{p} = h(\mathbf{z};\mathbf{S},\tau)$ and $p_i$ is the $i$-th item in $\mathbf{p}$.

\section{Method}

\label{sec:UCNS}
In this work, we aim to solve the challenging UniDA problem with Vision-Language Models (VLMs). 
As mentioned in the preliminary, we lack any prior knowledge about $\mathcal{C}_t$ and $\mathcal{T}^t$, resulting in uncertainty about the number of private classes in both domains.
Empirical evidence~(\citeauthor{unida_UAN,unida_CMU,saito2018open,cao2018partial}) suggests that sub-optimal domain alignment will emerge when target-common or target-private classes are mistakenly aligned with source-private or source-common classes, respectively.
Therefore, the primary challenge of UniDA lies in common class detection and the subsequent domain alignment.

\subsection{Existing Clustering-based Methods}
Recently, numerous target domain clustering-based UniDA methods~\cite{unida_DCC,unida_UniOT,unida_GLC} have been proposed and achieved promising results.
In this paper, we consider that the key prerequisite for the effectiveness of these methods can be summarized as accurately representing the semantics of both domains:

\begin{itemize}
	\item DCC~\cite{unida_DCC}: DCC utilizes source labels to compute source semantic centers and employs K-means centroids as target target semantic centers. Subsequently, common class matching is performed by cycle-consistent matching which is based on representation similarity. 
    \item GLC~\cite{unida_GLC}: For each class, in the target domain, GLC acquires positive prototypes using prediction results and negative prototypes via K-means. Then, GLC identifies common class samples based on similarity between samples and prototypes.
    \item UniOT~\cite{unida_UniOT}: UniOT first calculates the similarity between target samples and source prototypes, and then solves the optimal transport (OT) problem to classify samples as either common or unknown. Moreover, UniOT also obtains target prototypes through OT within the target domain.
\end{itemize}

\begin{figure*}[ht]
	\centering
	\includegraphics[scale=0.7]{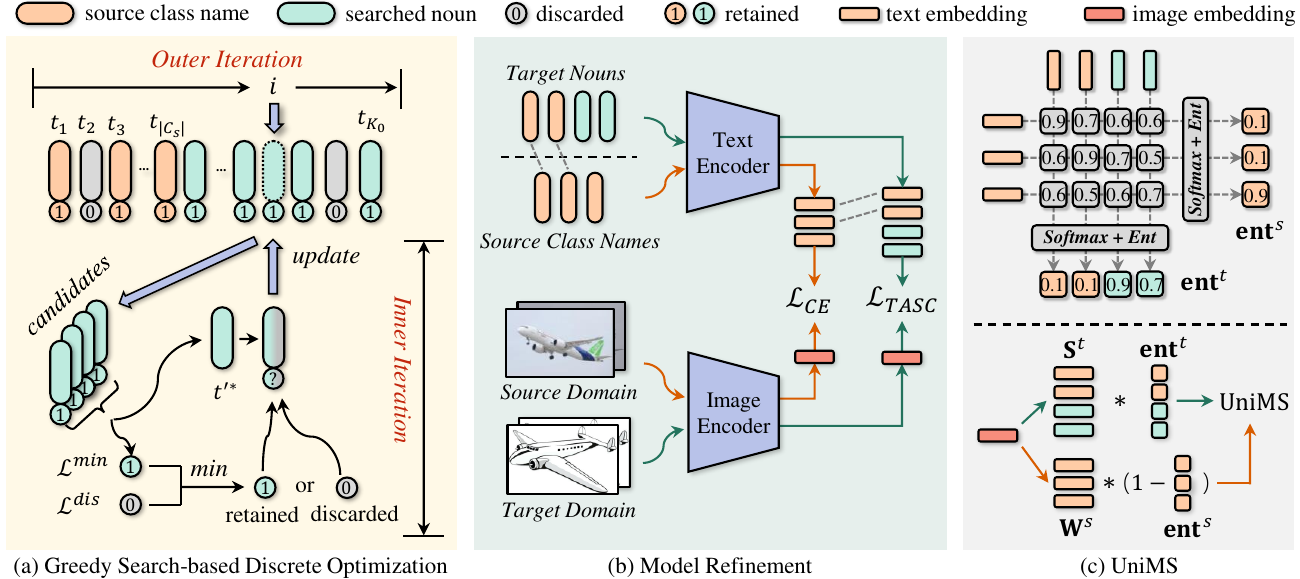}
	\caption{Overview of the proposed Target Semantics Clustering via Text Representations.}
	\label{fig:UCNS}
\end{figure*}

In summary, although these methods employ different techniques, they all aim to acquire semantic centers (\mbox{\ie}, prototypes or cluster centers) of both domains and detect common classes based on representation similarity.
However, the derived semantic centers are less than ideal, leading to complex and less robust algorithms. 
We argue that this is caused by two factors: 1) the derived semantic centers are domain-biased; 2) the number of classes in target domain is unknown.
These factors reduce the reliability of similarity and make it difficult to determine which domain contains private classes and how many such classes exist.
Unfortunately, current methods are powerless against them.

In this paper, we argue that this difficulty largely stems from the use of an unconstrained continuous image representation space in the process of representing semantics. 
First, these semantic centers are iteratively updated based on image representations in an unconstrained space lacking domain-invariant regularization, leading to the incorporation of domain-specific information.
Moreover, the points in this space are almost semantically meaningless and can possess any level of semantic granularity. 
Consequently, regardless of the number of clusters, the clustering loss can be effectively minimized, leading to the inability to estimate the number of clusters.

Based on these analyses, a straightforward idea emerges: constraining the search space of semantic centers to a semantically meaningful and less domain-biased representation space to facilitate simple and robust UniDA algorithm.

\subsection{Target Semantics Clustering (TASC)}
In this section, we will provide a concrete implementation of this idea.
Since the embeddings of $\mathcal{T}^s$ are sufficient to serve as the source semantic centers, we focus on the target domain. 
We first propose the mathematical formulation of Target Semantics Clustering (TASC) and then provide a two-stage optimization method, as shown in Figure~\ref{fig:UCNS}.

\subsubsection{Mathematical Formulation.} 
Benefiting from training on web-scale image-text pairs, text representations encoded by CLIP are inherently semantically meaningful and exhibit less domain bias.
Therefore, we leverage the embeddings of source class names $\mathcal{T}^{s}=(t_1^s, t_2^s, \dots, t_{|\mathcal{C}_s|}^s)$ and all nouns from WordNet~\cite{miller1995wordnet} $\mathcal{T}^{nouns}=(t_1^n, t_2^n, \dots, t_{N}^n)$ to construct the search space, where $N$ is the number of nouns.
Base on this, we can formulate the target domain clustering as the process of finding the optimal set of nouns $\mathcal{T}$ and the optimal model parameters $\theta$ to minimize $\mathcal{L}_{clu}$, \mbox{i.e.},
\begin{equation}
\begin{array}{l@{\quad}c@{\quad}l}
\min\limits_{\mathcal{T},\theta}  &  \mathcal{L}_{clu}(\mathcal{T}, \theta\,; \mathcal{D}^t) & \\[2ex]
\;\text{s.t.} & \mathcal{T}=(t_1, t_2, \dots, t_{K})  &\\[1ex]
            & t_i \in \mathcal{T}^s \cup \mathcal{T}^{nouns} \quad \forall\; i=1, 2, \dots, K
\end{array}
\end{equation}
where $K$ is the number of clusters. 
During optimization, the embeddings of $\mathcal{T}$ will serve as target semantic centers.

Although the above formulation constrains the search space of semantic centers, it relies on a predefined number of clusters $K$. To enable the optimization of the number of clusters, we additionally introduce a hidden state vector as: $\mathbf{r}=[r_1, r_2, \dots, r_{K_0}] \in \{0, 1\}^{K_0}$, 
where $K_0$ serves as the fixed upper bound of $K$.
The value of $r_i$ represents the status of $t_i$, \mbox{i.e.}, whether $t_i$ participates in the clustering: if $r_i=1$, $t_i$ is retained; if $r_i=0$, $t_i$ is discarded. 
Therefore, $K$ is dynamically determined by $\mathbf{r}$, \mbox{i.e.}, $K=\sum_{i=1}^{K_0} r_i$.
Finally, the TArget Semantics Clustering (TASC) via Text Representations can be formulated as:
\begin{equation}
\label{eqn:tasc}
\begin{array}{l@{\quad}c@{\quad}l}
\min\limits_{\mathcal{T}, \mathbf{r},\theta}  &  \mathcal{L}_{TASC}(\mathcal{T},  \mathbf{r}, \theta\,; \mathcal{D}^t) \triangleq \mathcal{L}_{clu}(\mathcal{T}^\mathbf{r}, \theta\,; \mathcal{D}^t)  &\\[2ex]
\;\text{s.t.} & \mathcal{T}=(t_1, t_2, \dots, t_{K_0})  &\\[1ex]
            & t_i \in \mathcal{T}^s \cup \mathcal{T}^{nouns} \quad \forall\; i=1, 2, \dots, K_0 &\\[1ex]
            & \mathbf{r}=[r_1, r_2, \dots, r_{K_0}]^T \in \{0, 1\}^{K_0} &\\[1ex]
            & \mathcal{T}^\mathbf{r} = (t_i|r_i=1, i=1, 2, \dots, K_0)
            
\end{array}
\end{equation}
where $\mathcal{T}^\mathbf{r}$ consists of all retained nouns from $\mathcal{T}$ based on $\mathbf{r}$. The text embeddings of $\mathcal{T}^{\mathbf{r}}$ will serve as the target domain semantic centers, denoted as $\mathbf{S}^t=[\mathbf{s}^t_1, \mathbf{s}^t_2, \dots, \mathbf{s}^t_K]$.

In terms of the loss function, we leave the explicit functional form of $\mathcal{L}_{clu}$ unspecified in the above discussion. 
This demonstrates that TASC is essentially an optimization framework that constrains the search space of clustering centers and enables estimating of the number of clusters.
In this paper, inspired by numerous works in source-free domain adaptation~(\citeauthor{sfda_shot,sfda_crco}), we adopt information maximization as the objective for clustering. Using the notations in \mbox{Eq. (\ref{eqn:func_ent})}, we instantiate $\mathcal{L}_{clu}$ as:
\begin{equation}
\label{eqn:loss_clu}
\begin{aligned}
    \mathcal{L}_{clu} &= \mathcal{L}_{ent} + \lambda_{div}\mathcal{L}_{div},\\
    &= \mathbb{E}_{\mathbf{x}^t\in \mathcal{D}^t} \;\text{Entropy}(\mathbf{p}(\mathbf{x}^t))
     -\lambda_{div}\text{Entropy} (\bar{\mathbf{p}}),
\end{aligned}
\end{equation}
where $\mathbf{p}(\mathbf{x}^t) = h(f(\mathbf{x}^t);\mathbf{S}^t, \tau)$ is the prediction score of $\mathbf{x}^t$, $\bar{\mathbf{p}}=\mathbb{E}_{\mathbf{x}^t\in \mathcal{D}^t} \mathbf{{p}(\mathbf{x}^t)}$ is the mean prediction score of all samples, and $\lambda_{div}$ is a trade-off hyperparameter.

We have now successfully formulated the Target Semantics Clustering via Text Representations. 
However, as evident from \mbox{Eq.~(\ref{eqn:tasc})} and \mbox{Eq.~(\ref{eqn:loss_clu})}, this is an extremely challenging Mixed-Integer Nonlinear Programming (MINLP)~\cite{belotti2013mixed} problem.
To address this, in this paper, we offer a practicable solution that consists of two stages.

\subsubsection{Greedy Search-based Discrete Optimization.}
In this optimization stage, the model parameters are fixed, and only the discrete variables, \mbox{i.e.}, $\mathcal{T}$ and $\mathbf{r}$, are optimized. During initialization, $\mathbf{r}$ is set to an all-ones vector. For $\mathcal{T}$, we place $\mathcal{T}^s$ at the front part of it and randomly initialize the remaining $K_0-|\mathcal{C}_s|$ nouns, \mbox{i.e.},
\begin{equation}
\label{eqn:T_init}
    \mathcal{T}=(t_1^s, t_2^s, \dots, t_{|\mathcal{C}_s|}^s, t_{|\mathcal{C}_s|+1}, \dots, t_{K_0}).
\end{equation}

Due to the excessively large search space of $\mathcal{T}$ and $\mathbf{r}$, we employ a greedy optimization strategy. Specifically, at the \mbox{$i$-th} step, we optimize only the noun $t_i$ and its status $r_i$ while keeping the other variables in $\mathcal{T}$ and $\mathbf{r}$ fixed. 
Regarding a single step, we provide the following summary and analysis:
\begin{enumerate}
    \item First, construct the feasible space to be searched. For $t_i$, randomly select $n_c$ candidates $\mathcal{T}^{c}$ from $\mathcal{T}^{nouns}$; for $r_i$, it is binary, being either 0 or 1. Moreover, when $r_i=0$, $t_i$ will not participate in calculating $\mathcal{L}_{TACS}$ since it is discarded. Therefore, we only need to explore $(n_c+1)$ possible solutions. Let use $\mathcal{T}_{i|t^\prime}$ to denote that the $i$-th item of $\mathcal{T}$ is replaced by $t^\prime$ and similarly for $\mathbf{r}_{i|0}$ and $\mathbf{r}_{i|1}$.
    \item Then, find the optimal $t^{\prime*}$ from the $n_c$ candidates when $r_i=1$, as follows:
    \begin{gather}
        \mathcal{L}^{min} = \min\limits_{t^\prime \in \mathcal{T}^c} \mathcal{L}_{TASC}(\mathcal{T}_{i|t^\prime},  \mathbf{r}_{i|1}, \theta\,; \mathcal{D}^t).
    \end{gather}
    The loss when $r_i=0$ is calculated as:
    \begin{gather}
        \mathcal{L}^{dis} = \mathcal{L}_{TASC}(\mathcal{T},  \mathbf{r}_{i|0}, \theta\,; \mathcal{D}^t).
    \end{gather}
    \item Finally, update the discrete variables. Use $\mathcal{T}_{i|t^{\prime*}}$ to update $\mathcal{T}$. Set $r_i$ to 1 if $\mathcal{L}^{min} < \mathcal{L}^{dis}$; otherwise, set it to 0.
\end{enumerate}

Based on the single step update, we traverse i from 1 to $K_0$ to perform a total of $K_0$ steps, constituting one outer iteration.
Throughout the entire discrete optimization, we will conduct $N_{outer}$ outer iterations.
To enhance clarity and rigor, we've summarized the above process in Appendix.

It is worth mentioning that, from the perspective of domain adaptation, we should ensure that the semantic centers of common classes across the two domains are identical or highly similar, so that the target domain clustering can substantially contribute to enhancing the classifier's performance. 
To this end, for the first $|\mathcal{C}_s|$ items in $\mathcal{T}$ and $\mathbf{r}$, we employ two dedicated designs: 1) Starting from the initialization, never update the first $|\mathcal{C}_s|$ items in $\mathcal{T}$. 2) Obtain target domain prototypes $\boldsymbol{\mu}$ using the current predictions based on $\mathcal{T}^{\mathbf{r}}$. When $\text{Entropy}(h(g(t_i^s);\boldsymbol{\mu},\tau)) < \gamma_{ent}$, set $r_i$ to 1; otherwise, update $r_i$ based on $\mathcal{L}_{min}$ and $\mathcal{L}_{dis}$ as normal.

After this stage, we obtain the optimal $\mathcal{T}^*$ and $\mathbf{r}^*$.  
Due to the dedicated designs, the source-private classes initially included in $\mathcal{T}$ will be adaptively discarded, while the common classes are retained, and the unknown target-private classes will be represented by the searched nouns. 

\subsubsection{Model Refinement.}
In this optimization stage, the discrete variables ($\mathcal{T}^*$ and $\mathbf{r}^*$) are fixed, and only the model parameters $\theta$ are optimized.
Consequently, based on \mbox{Eq. (\ref{eqn:tasc})}, TASC reduces to a standard neural network optimization problem that can be addressed via gradient descent:
\begin{equation}
\label{eqn:refine}
\min\limits_{\theta} \quad\mathcal{L}_{TASC}(\mathcal{T}^*,  \mathbf{r}^*, \theta\,; \mathcal{D}^t) = \mathcal{L}_{clu}(\mathcal{T}^{*\mathbf{r}^*}, \theta\,; \mathcal{D}^t).
\end{equation}
Moreover, consistent with existing DA methods, we utilize the Cross Entropy loss on source domain to guide the adaptation of the generic VLMs to the target task.
In summary, we adopt the following losses for model refinement:
\begin{equation}
\label{eqn:loss_all}
\begin{aligned}
\min\limits_{\theta} \quad \mathcal{L}_{all} = \mathcal{L}_{CE} + \mathcal{L}_{TASC}.
\end{aligned}
\end{equation}

It is worth noting that, we calculate $\mathcal{L}_{CE}$ using the source domain semantic centers $\mathbf{W}^s$ obtained from $\mathcal{T}^s$; whereas in $\mathcal{L}_{TASC}$, we compute it using the target domain semantic centers $\mathbf{S}^t$ derived from $\mathcal{T}^{*\mathbf{r}^*}$.
Benefiting from the the dedicated designs in the discrete optimization, the semantic centers of common classes in $\mathbf{W}^s$ and $\mathbf{S}^t$ are nearly identical, enabling $\mathcal{L}_{TASC}$ achieves common class alignment and private class clustering simultaneously. 
Additionally, from the perspective of loss design, we employ the information maximization criterion derived from Closed-set DA, achieving simple and robust Universal DA, without explicitly identifying target-private samples during optimization.

\begin{table*}[h]
  \begin{center}
  
  \setlength{\tabcolsep}{1mm}
  \small
  \begin{tabular}{l|c||cccc|c|cccc|c||cccccc|c|c}
  \toprule
  \multirow{2}{*}{Method} & &
  \multicolumn{5}{c|}{\textbf{Office}} &
  \multicolumn{5}{c||}{\textbf{Office-Home}} &
    \multicolumn{7}{c|}{\textbf{DomainNet}} &
  \multicolumn{1}{c}{\textbf{VisDA}}
  \\
  \cmidrule(r){3-20}
   & &
    \multicolumn{1}{c}{OPDA} &
    \multicolumn{1}{c}{ODA} &
    \multicolumn{1}{c}{PDA} &
    \multicolumn{1}{c|}{CDA} &
    \multicolumn{1}{c|}{\textbf{Avg}} &
    \multicolumn{1}{c}{OPDA} &
    \multicolumn{1}{c}{ODA} &
    \multicolumn{1}{c}{PDA} &
    \multicolumn{1}{c|}{CDA} &
    \multicolumn{1}{c||}{\textbf{Avg}} &
    \multicolumn{1}{c}{PR} &
    \multicolumn{1}{c}{PS} &
    \multicolumn{1}{c}{RP} &
    \multicolumn{1}{c}{RS} &
    \multicolumn{1}{c}{SP} &
    \multicolumn{1}{c|}{SR} &
    \multicolumn{1}{c|}{\textbf{Avg}} & 
    \multicolumn{1}{c}{SR}
    \\  
  \midrule
    SO &  \multirow{7}{*}{\rotatebox{90}{ResNet50}} & 64.9 & 69.6 & 87.8 & - & - & 60.9 & 55.2 & 62.9 & - & - & 57.3 & 38.2 & 47.8 & 38.4 & 32.2 & 48.2 & 43.7 & 25.7\\
    DCC & & 80.2 & 72.7 & 93.3 & - & - & 70.2 & 61.7 & 70.9 & - & - & 56.9 & 43.7 & 50.3 & 43.3 & 44.9 & 56.2 & 49.2 & 43.0\\
    DANCE & & 80.3 & 79.8 & 86.0 & - & - & 49.2 & 12.9 & 71.1 & - & - & 21.0 & 37.0 & 47.3 & 46.7 & 27.7 & 21.0 & 33.5 & 42.8 \\
    OVANet & & 86.5 & 91.7 & 74.6 & - & - & 71.8 & 64.0 & 49.5 & - & - & 56.0 & 47.1 & 51.7 & 44.9 & 47.4 & 57.2 & 50.7 & 53.1\\
    GATE & & 87.6 & 89.5 & 93.7 & - & - & 75.6 & 69.0 & 74.0 & - & - & 57.4 & 48.7 & 52.8 & 47.6 & 49.5 & 56.3 & 52.1 & 56.4\\
    GLC & & 87.8 & 89.0 & 94.1 & - & - & 75.6 & 69.8 & 72.5& - & - & 63.3 & 50.5 & 54.9 & 50.9 & 49.6 & 61.3 & 55.1 & 73.1\\
    UniOT & & 91.1 & - & - & - & - & 76.6 & - & - & - & - & 59.3 & 51.8 & 47.8 & 48.3 & 46.8 & 58.3 & 52.0 & 57.3\\
    \midrule
    SO$^{\star}$ & \multirow{5}{*}{\rotatebox{90}{CLIP}} & 75.8 & 83.5 & 95.1 & 87.3  & 85.4 & 76.2 & 64.6 & 81.8 & 79.8 & 75.6 & 67.1 & 63.0 & 64.0 & 63.2 & 56.9 & 66.9 & 63.5 & 51.8 \\
    DANCE$^{\star}$ & & 89.7 & 90.7 & 88.1 & 87.5  & 89.0 & 83.9 & 65.4 & \underline{84.6} & 81.7 & 78.9 & 66.0 & 62.1 & 65.3 & 63.9 & \underline{62.6} & 66.4 & 64.4 & 75.3 \\
    UniOT$^{\star}$ & & 91.1 & 93.7 & 71.2 & \underline{88.9}  & 86.2 & \underline{86.3} & \underline{80.3} & 74.8 & \underline{82.1} & \underline{80.9} & 72.0 & 63.8 & \underline{66.5} & \underline{66.5} & 62.0 & \underline{72.8} & \underline{67.3} & 75.3 \\
    OVANet$^{\star}$ & & \textbf{92.1} & \underline{94.5} & \underline{95.2} & 86.7  & \underline{92.1} & 85.0 & 76.2 & 81.9 & 80.0 & 80.8 & \underline{72.5} & \underline{64.5} & 65.6 & 65.2 & 61.8 & 72.1 & 67.0 & \underline{79.1} \\
    Ours & & \underline{91.3} & \textbf{96.1} & \textbf{96.3} & \textbf{90.1}  & \textbf{93.5}  & \textbf{89.4} & \textbf{83.8} & \textbf{89.2} & \textbf{86.7} & \textbf{85.4} & \textbf{80.5} & \textbf{69.7} & \textbf{69.1} & \textbf{69.3} & \textbf{69.6} & \textbf{81.5} & \textbf{73.3} & \textbf{90.4}\\
    \bottomrule
    \end{tabular}
  \caption{Performance comparison between state-of-the-arts and our method. The left part shows the results on Office (average on 6 tasks) and Office-Home (average on 12 tasks) under four different category shift scenarios, and the right part shows the results on large-scale DomainNet and VisDA under OPDA scenarios. We report the H-score for OPDA and ODA, as well as the classification accuracy for PDA and CDA. Some results are referred to previous work~\cite{unida_GLC}.}
  \label{table:office_officehome}
  \end{center}
\end{table*}

\subsection{Universal Maximum Similarity}
\label{sssec:UniMLS}
During inference, we need to assign target-private samples to a single \textit{unknown} class.
For unknown detection, existing methods often construct a scoring function based on entropy~(\citeauthor{unida_UAN,unida_CMU,unida_DANCE}), confidence~\cite{unida_CMU}, similarity~(\citeauthor{unida_DCC,unida_UniOT}), etc., with a manual threshold for the final decision.
However, they lack robustness across different category shift scenarios due to their inability to accurately perceive category shifts.
In this paper, based on the high-quality semantic centers, we first explicitly model the category shift and then embed it into the similarity score for more robust unknown detection.

\begin{table}[t]
  \begin{center}
  \setlength{\tabcolsep}{1.4mm}

  \small
  \begin{tabular}{ccc|cccc|c}

    \toprule
    \multirow{2}{*}{$\mathcal{L}_{CE}$} & \multirow{2}{*}{$\mathcal{L}_{TASC}$} & \multirow{2}{*}{UniMS}& \multicolumn{4}{c|}{\textbf{Office}}  & \multirow{2}{*}{\textbf{Avg}}\\
    \cmidrule(r){4-7}
    & & & OPDA & ODA & PDA & CDA \\
    \midrule
    \cmark &        &        & 82.9  & 90.9  & \underline{95.1}  & \underline{87.3}  & 89.1  \\
    \cmark &        & \cmark & \underline{89.6}  & \underline{95.2}  & \underline{95.1}  & \underline{87.3}  & \underline{91.8} \\
    \cmark & \cmark & \cmark & \textbf{91.3} & \textbf{96.1} & \textbf{96.3} & \textbf{90.1} & \textbf{93.5} \\

    \bottomrule
    \end{tabular}
  \caption{Ablation studies on training loss and scoring function. Experiments conduct on Office (average on 6 tasks).}
    \label{table:ablations_loss}
  \end{center}
  \end{table}

First, functionally speaking, both ${\mathbf{S}^t}$ and $\mathbf{W}^s$ can serve as image classifiers. 
More broadly, we can utilize them to classify for each other.
Based on the classification scores, the following two entropy vectors can be further defined: 
\begin{align}
\mathbf{ent}^s &=[ent_1^s, ent_2^s, \dots, ent_{|\mathcal{C}_s|}^s]^T \in [0, 1]^{|\mathcal{C}_s|}, \\
\mathbf{ent}^t &=[ent_1^t, ent_2^t, \dots, ent_K^t]^T  \in [0, 1]^{K},
\end{align}
where $ent_i^s = \text{Entropy}(h(\mathbf{w}_i^s;\mathbf{S}^t, \tau)) / \log{K}$ and $ent_j^t = \text{Entropy}(h(\mathbf{s}_j^t;\mathbf{W}^s, \tau)) / \log{|\mathcal{C}_s|}$ are the normalized entropy that bounded in the range of $[0, 1]$.
Due to the constrained space and dedicated designs, the representation similarity is reliable enough for $\mathbf{ent}^s$ and $\mathbf{ent}^t$ to perceive category shifts.
In other words, lower values of ${ent}^s_i$ and ${ent}^t_j$ indicate a higher probability that $\mathbf{w}_i^s$ and $\mathbf{s}_j^t$ correspond to common classes.
To leverage this information, we embed both $\mathbf{ent}^s$ and $\mathbf{ent}^t$ into the widely adopted MLS~\cite{vaze2021open} to derive the Universal Maximum Similarity (UniMS) as:
\begin{equation}
\label{eqn:UniMS}
\begin{aligned}
    \text{UniMS}(\mathbf{x}^t) = &\max \{(1-ent^s_i) * \text{sim}(f(\mathbf{x}^t), \mathbf{w}^s_i)\}_{i=1}^{\lvert\mathcal{C}_s\rvert}\\
    &- \max \{ent^t_j * \text{sim}(f(\mathbf{x}^t), \mathbf{s}^t_j)\}_{j=1}^K.
\end{aligned}
\end{equation}
Intuitively, if an image embedding $f(\mathbf{x}^t)$ is close to $\mathbf{w}^s_i$ and $ent^s_i$ is near $0$, it will obtain a higher UniMS score; conversely, if it is close to $\mathbf{s}^t_j$ and $ent^t_j$ is near $1$, the UniMS score will be suppressed.

Inspired by~(\citeauthor{jahan2024unknown,jang2022unknown}), we optimize a 2-component Gaussian Mixture Model (GMM) to obtain the adaptive threshold of UniMS for unknown detection. 
During optimization, unlike existing works, we utilize the proportion of target-private classes estimated by TASC and fix it as the mixture weights. 
Specifically, the number of target-private classes can be estimated via $\sum_{i=|\mathcal{C}_s|+1}^{K_0}r_i$. 
During prediction, we set the mixture weights to be uniform to achieve theoretically optimal results. More details and theoretical proofs are provided in the Appendix.

\begin{table}[t]
  \begin{center}

  \small
  \begin{tabular}{c|cc|cc|c}
    \toprule
    \multirow{2}{*}{\textbf{Weights}} & \multicolumn{2}{c|}{\textbf{Office}} & \multicolumn{2}{c|}{\textbf{Office-Home}} & \multirow{2}{*}{\textbf{Avg}}\\
    \cmidrule(r){2-3}\cmidrule(r){4-5} 
    & OPDA & ODA & OPDA & ODA \\
    \midrule
    \xmark & 91.0 & \textbf{96.3} & 84.9 & 81.9 & 88.5 \\
    \cmark & \textbf{91.3} & 96.1 & \textbf{89.4} & \textbf{83.8} & \textbf{90.2} \\
    
    \bottomrule
    \end{tabular}
  \caption{H-score (\%) on Office-Home (average on 12 tasks) and Office (average on 6 tasks) under OPDA and ODA. ``Weights'' indicates whether the estimated proportion of target-domain private classes is used in GMM optimization.}
    \label{table:GMM_Weights}
  \end{center}
  \end{table}

\section{Experiments}
\label{sec:exp}

\subsection{Setup}
\noindent\textbf{Dataset.}
Our method will be validated on four popular datasets in Domain Adaptation, \ie, 
Office~\cite{office}, Office-Home~(\citeauthor{officehome}), VisDA~\cite{visda}, and DomainNet~\cite{domainnet}. 
Due to the large amount of data, we only conduct experiments on three subsets from DomainNet, \ie, Painting(P), Real(R), and Sketch(S), following existing works~(\citeauthor{unida_UniOT,unida_OVANet,unida_GLC}). 
We evaluate our method on 4 different category shift scenarios to demonstrate its robustness, \ie, CDA, PDA, ODA, and OPDA. 
Detailed classes split in these scenarios are summarized in Appendix, which is the same as dataset split in~\cite{unida_GLC}.

\noindent\textbf{Implementation Details.}
We use pre-trained CLIP model with ViT-B/16~(\citeauthor{dosovitskiy2020image}) and Transformer~(\citeauthor{vaswani2017attention}) as image and text encoders, respectively.
LoRA~\cite{hu2021lora} is used in all transformer blocks in both image and text encoders with $rank=8$.
More details about LoRA can be found in the Appendix. 
We adopt the same learning rate scheduler $\eta=\eta_0 \cdot (1+10\cdot p)^{-0.75}$ as~(\citeauthor{long2018conditional,sfda_shot}), where $p$ is the training progress changing from 0 to 1 and $\eta_0=0.0001$.
For the hyper-parameters, we empirically set the $\lambda_{div}$ to 0.6, which differs from SHOT-IM~(\citeauthor{sfda_shot}), and we will discuss this in details later.
$\tau$ is set to 0.02.
In the discrete optimization step of TASC, $n_c=300$, $\gamma_{ent}=0.3$, and $N_{outer}=20$.
$K_0$ is set to 100 for Office, Office-Home, and VisDA, but 400 for DomainNet.

\noindent\textbf{Evaluation Protocols.}
For a fair comparison, we follow the same evaluation metric as previous works~(\citeauthor{unida_GLC,unida_COCA,unida_CMU,unida_UniOT,unida_DCC}).
In OPDA and ODA scenarios, we report the H-score~\cite{unida_CMU}. 
The \mbox{H-score}, as defined by~\cite{unida_CMU}, is the harmonic mean of the accuracy $a_{\mathcal{C}}$ on common classes and the accuracy $a_{\bar{\mathcal{C}}_t}$ on a single unknown class. 
In PDA and CDA scenarios, we report the classification accuracy over all target samples following~\cite{unida_DCC,unida_GLC}.

\noindent\textbf{Compared Methods.}
We select multiple state-of-the-art methods for comparison, including DCC~\cite{unida_DCC}, DANCE~(\citeauthor{unida_DANCE}), OVANet~(\citeauthor{unida_OVANet}), GLC~(\citeauthor{unida_GLC}), UniOT~(\citeauthor{unida_UniOT}). 
Additionally, we utilize CLIP~\cite{CLIP} as backbone and fine-tune it via LoRA~\cite{hu2021lora}, while most existing state-of-the-arts fully fine-tune ResNet-50 or ViT-B/16 pre-trained on ImageNet~\cite{imagenet}.
Therefore, for a fair comparison, we conduct experiments on following methods under the same conditions as ours: Source-Only (SO), DANCE, OVANet, UniOT, marked as $^{\star}$.

\begin{table}[t]

  \begin{center}
  \small

  \begin{tabular}{l|cc|cc|c}
    \toprule
    \multirow{2}{*}{\textbf{Method}} & \multicolumn{2}{c|}{\textbf{Office}} & \multicolumn{2}{c|}{\textbf{Office-Home}} & \multirow{2}{*}{\textbf{Avg}}\\
    \cmidrule(r){2-5}
    & OPDA & ODA & OPDA & ODA \\
    \midrule
    MS-t & 58.4 & 73.3 & 59.4 & 51.4 & 60.6 \\
    MS-t w/ $\mathbf{ent}^t$ & 90.3 & 91.0 & 84.2 & 75.5 & 85.2 \\
    MS-s & 93.1 & \underline{99.0} & 95.1 & \underline{92.2} & 94.8 \\
    MS-s w/ $\mathbf{ent}^s$ & \underline{93.4} & \underline{99.0} & \underline{95.9} & 92.0 & \underline{95.1} \\
    UniMS & \textbf{96.6} & \textbf{99.1} & \textbf{96.4} & \textbf{92.6} & \textbf{96.2} \\
    
    \bottomrule
    \end{tabular}
  \caption{AUROC (\%) on Office-Home (average on 12 tasks) and Office (average on 6 tasks) under OPDA and ODA. Let denote the first item of UniMS as MS-s w/ $\mathbf{ent}^s$ and its unweighted version as MS-s; similarly for the second item.}
    \label{table:AUROC}
  \end{center}
  \end{table}

\subsection{Results}

\noindent\textbf{Comparison with state-of-the-arts.}
As shown in Table~\ref{table:office_officehome}, our method achieves the best average performance on all the four benchmarks without tuning any hyper-parameters (except $K_0$). 
Notably outstanding is that we exceed existing methods by considerable margins of 6.0\% and 11.3\% on the large-scale DomainNet and VisDA, respectively. 
All these results demonstrates that our approach is quite effective and more robust when addressing different category shifts.

\noindent\textbf{Ablation studies of TASC.}
For all experiments in Table~\ref{table:ablations_loss}, we perform the discrete optimization stage of TASC, followed by the selective use of UniMS and $\mathcal{L}_{TASC}$ in the model refinement stage. 
The consistent improvements demonstrate their effectiveness.

\begin{figure}[!t]
\centering
\includegraphics[width=0.85\columnwidth]{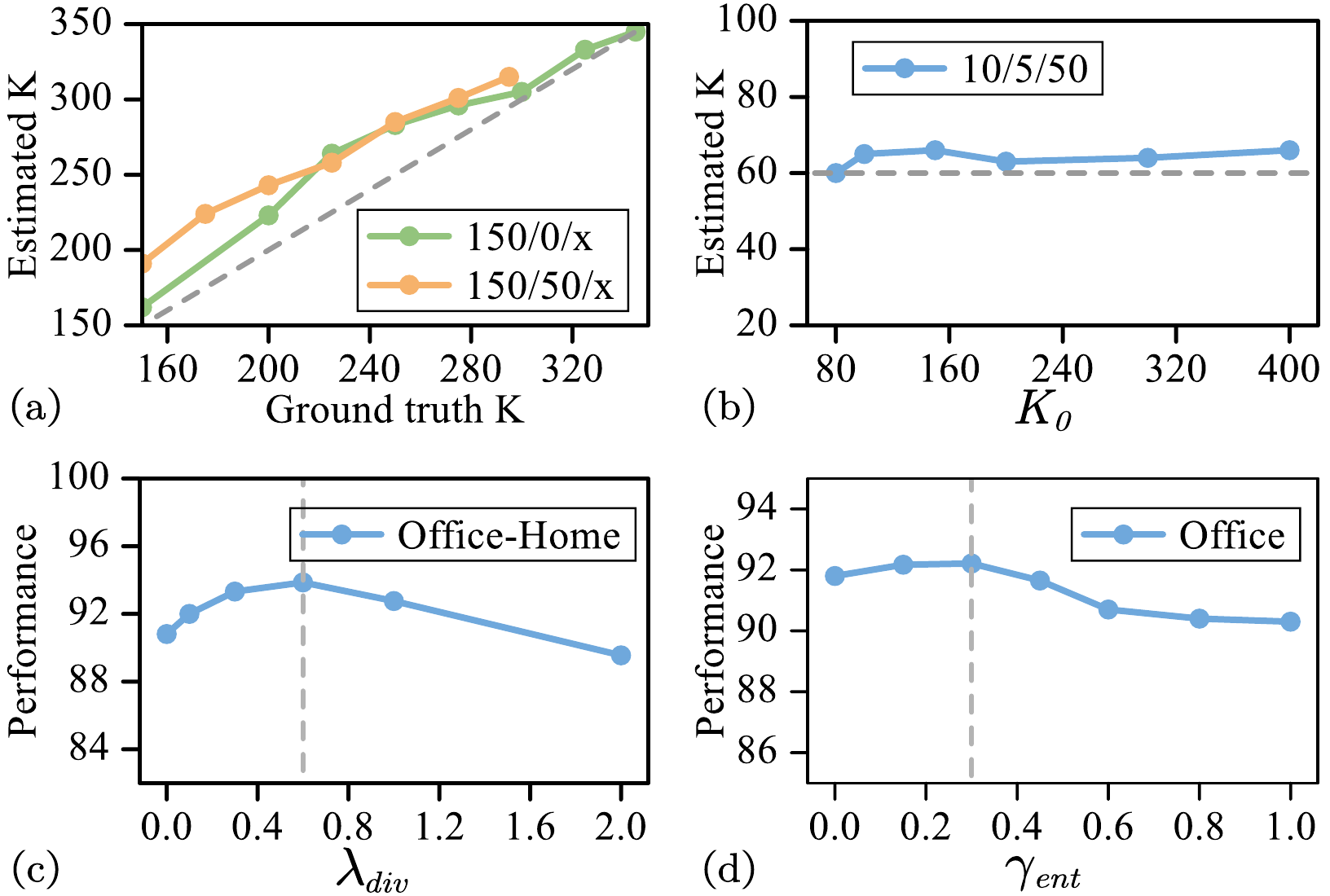}

\caption{(a) Effectiveness of the adaptive estimation of $K$. (b) Sensitivity of $K_0$. (Office-Home, OPDA, AC) (c) Sensitivity of $\lambda_{div}$. 
(Officehome, OPDA, average on AC, CP, PR, RA). (d) Sensitivity of $\gamma_{ent}$. (Office, OPDA, average on AD, DW, WA).}
\label{fig:other}
\end{figure}

\noindent\textbf{Estimation of the number of clusters.} To evaluate the adaptive estimation mechanism, we vary the dataset split and plot the estimated $K$ in a line chart. 
Let use "$|\mathcal{C}|/|\bar{\mathcal{C}}_s|/|\bar{\mathcal{C}}_t|$" to denote different category shifts. In Figure~\ref{fig:other}(a), we consider two settings: ``$150/0/x$'' and ``$150/50/x$'', on SR of DomainNet, where $x$ represents the varying numbers and $K_0=400$.
As shown by the two color lines, although with the same initial $K_0$, when varying the number of categories, our algorithm consistently converges towards the ground truth number adaptively.

\noindent\textbf{Effectiveness of perceiving category shifts.}
First, we conduct ablation studies on the different components of UniMS based on AUROC metric.
As shown in \mbox{Table~\ref{table:AUROC}}, MS-s serves as a good baseline. Moreover, the category shift information perceived by $\mathbf{ent}^t$ and $\mathbf{ent}^s$ enhances MS-t and MS-s. Finally, MS-t w/ $\mathbf{ent}^t$ provides complementary information to MS-s w/ $\mathbf{ent}^s$, enabling UniMS to achieve the highest performance.
Second, we evaluate the effectiveness of setting GMM mixture weights to the estimated proportion of target-private classes by TASC. Table~\ref{table:GMM_Weights} demonstrates that using this proportion is more effective, especially for severely imbalanced category shifts like Office-Home under OPDA (10/5/50) and ODA (25/0/40).

\noindent\textbf{Parameter sensitivity.}
Figure~\ref{fig:other} (b-d) presents our sensitivity analysis on $\gamma_{ent}$, $K_0$, and $\lambda_{div}$, showing their stability within specific ranges. Notably, the optimal $\lambda_{div}$ is lower than 1.0 in SHOT-IM, likely due to partially undiscarded private classes and over-clustering in $\mathcal{T}^{\mathbf{r}}$, necessitating a reduction in the requirement for diversity.

\section{Conclusion}
In this paper, we propose that uniformly represent the image semantics of both domains in the semantically rich and discrete text representation space can facilitate simple and robust UniDA algorithm.
Based on this idea, we propose target semantics clustering via text representations and universal maximum similarity.
Extensive experiments demonstrate the robustness and effectiveness.
Finally, considering the universality of TASC, we expect the development of additional UniDA algorithms built upon this framework.

\section{Acknowledgments}
This work is supported by the National Natural Science Foundation of China under Grant 62176246. This work is also supported by Anhui Province Key Research and Development Plan (202304a05020045) and Anhui Province Natural Science Foundation (2208085UD17). This work is also supported by National Natural Science Foundation of China under Grant 62406098 and 62376256, and The Joint Fund for Medical Artificial Intelligence under Grant MAI2022Q011.

\bibliography{aaai25}

\newpage

\appendix

\twocolumn[
  \begin{center}
    {\LARGE \bfseries Appendix:

Target Semantics Clustering via Text Representations

for Robust Universal Domain Adaptation \par}

\vspace{2em}
  \end{center}
]

\setcounter{secnumdepth}{1} 
\renewcommand{\thesection}{\Alph{section}}
\renewcommand{\thefigure}{A\arabic{figure}}
\setcounter{figure}{0}
\renewcommand{\thetable}{A\arabic{table}}
\setcounter{table}{0}
\setcounter{equation}{15}

\section{LoRA Fine-tuning}
\label{sec:lora}

To efficiently transfer the CLIP model to the target domain, inspired by~\cite{smith2023construct, doveh2023teaching, cascante2023going}, we fine-tune both the image encoder $f$ and text encoder $g$ of CLIP models~\cite{CLIP} via LoRA~\cite{hu2021lora}. 
For any pre-trained weight matrix $W_0\in\mathbb{R}^{m \times n}$ of linear layer within Transformer~\cite{vaswani2017attention} architecture, LoRA constrains its update by representing the residual with a low-rank decomposition $W_0+\Delta W=W_0+BA$, where $B\in\mathbb{R}^{m\times r}$ and $A\in\mathbb{R}^{r\times n}$ are the residual weights, and the rank $r\ll \text{min}(m, n)$.
$W_0$ is frozen while $A$ and $B$ are trainable during training.
We follow the same initialization for $A$ and $B$ in~\cite{hu2021lora}, as well as simply set $\alpha$ to $r$.
Empirically, we set $r=8$ and only apply LoRA to $W_q$ and $W_v$ in each layer which are the weight matrices in the self-attention module.
During inference, we merge the residual weights $A$ and $B$ into $W_0$, \ie, $W = W_0 + BA$ for forward process.

\section{Details of UniMS and GMM}
\label{sec:scoring_func}
As mentioned in the main paper, we introduce a novel scoring function for target-private samples detection in UniDA, named Universal Maximum Similarity (UniMS). During inference, we optimize a Gaussian Mixture Model (GMM) to obtain the discriminative threshold. Here, we first provide the detailed definitions of various variants of UniMS used in Table 3 and then formulate the process for obtaining the discriminative threshold.

\subsection{Definition of Components in UniMS}
To facilitate the analysis of the effectiveness of each component in UniMS, we define the following scoring functions as:
\begin{align}
\small
\text{MS-t}\; (\mathbf{x}^t)&=-\max \{\text{sim}(f(\mathbf{x}^t), \mathbf{s}^t_j)\}_{j=1}^K, \notag \\ 
\small
\text{MS-s}\; (\mathbf{x}^t)&=\max \{\text{sim}(f(\mathbf{x}^t), \mathbf{w}^s_i)\}_{i=1}^{\lvert\mathcal{C}_s\rvert}, \notag \\
\small
\text{MS-t w/ $\mathbf{ent}^t$}\; (\mathbf{x}^t)&= -\max \{ent^t_j * \text{sim}(f(\mathbf{x}^t), \mathbf{s}^t_j)\}_{j=1}^K, \notag  \\
\small
\text{MS-s w/ $\mathbf{ent}^s$}\; (\mathbf{x}^t)&= \max \{(1-ent^s_i) * \text{sim}(f(\mathbf{x}^t), \mathbf{w}^s_i)\}_{i=1}^{\lvert\mathcal{C}_s\rvert}.  \notag 
\end{align}
Note that, MS-s is essentially equivalent to MLS~\cite{vaze2021open} when substituting the learned classifier weights of the latter with text embeddings. 
As evaluated in the main paper, with the weighting via $\mathbf{ent}^s$ and $\mathbf{ent}^t$, we eliminate the bad influence of category shift.
Besides, while the performance of MS-t w/ $\mathbf{ent}^t$ is far inferior to MS-s w/ $\mathbf{ent}^s$, it provides complementary information to \mbox{MS-s w/ $\mathbf{ent}^s$}, resulting in higher performance of the ensemble~(\ie, UniMS).

\subsection{GMM-based Optimal Threshold Selection}
To adaptively obtain the threshold based on UniMS, inspired by~\cite{jahan2024unknown,jang2022unknown}, we adopt a Gaussian Mixture Model (GMM). First, we formulate the GMM as follows:
\begin{gather}
    p(m|y=k) = \phi(m|\mu_k, \sigma_k), \\
    p(m|y=u) = \phi(m|\mu_u, \sigma_u), \\
    p(m) = p_k p(m|y=k) + p_u p(m|y=u), \\
    p(m|\theta_{GMM})= p_k\phi(m|\mu_k, \sigma_k) + p_u\phi(m|\mu_u, \sigma_u),
\end{gather}
where $\theta_{GMM}=\{(p_k, \mu_k, \sigma_k), (p_u,\mu_u, \sigma_u)\}$ is the parameters of GMM, $k$ and $u$ are the abbreviation of ``known'' and ``unknown'', $m=\text{UniMS}(\mathbf{x}^t)$ represents the UniMS score, and $(p_k, p_u)$ is the mixture weights.

Regarding the mixture weights, \cite{jahan2024unknown}~uses equal priors, \mbox{i.e.}, $(p_k, p_u)=(0.5, 0.5)$, as the mixture weights for modeling the GMM, while \cite{jang2022unknown}~employs the beta mixture model itself to estimate its mixture weights. In contrast, as mentioned in the main text, we use the proportion of unknown classes estimated by TASC as the mixture weights. Specifically, our GMM model can be formalized as:
\begin{gather}
\label{eqn:gmm_ours}
    (\hat{p}_k, \hat{p}_u) = (\frac{K_{com}}{K_{com}+K_{pri}}, \frac{K_{pri}}{K_{com}+K_{pri}}), \\
    \label{eqn:gmm}
    p(m|\theta_{GMM}^\prime)= \hat{p}_k\phi(m|\mu_k, \sigma_k) + \hat{p}_u\phi(m|\mu_u, \sigma_u),
\end{gather}
where $K_{com} = \sum_{i=1}^{|\mathcal{C}_s|} r_i$ and $K_{pri}=\sum_{i=|\mathcal{C}_s|}^{K_0} r_i$ are the estimated number of target-common classes and the estimated number of target-private classes, respectively.
It's important to note that accurate mixture weights positively affect the estimation of other GMM parameters, specifically the means and covariances.
Let denote the estimated means and covariances as $(\hat{\mu}_c, \hat{\sigma}_c)\; \forall \; c={k, u}$.

On the other hand, during inference, we reset the mixture weights of the GMM to be uniform, i.e., $(p_k, p_u) = (0.5, 0.5)$, a strategy based on theoretical reasoning. Specifically, we have a detailed formulation to explain and justify this strategy. 
First, in the UniDA task, we should balance the accuracy of both known and unknown classes~\cite{unida_CMU}. To this end, we can formalize the objective as:
\begin{align}
   \min \quad L = \frac{1}{2} \sum_{c={k, u}} \mathbb{E}_{y(\mathbf{x}^t)=c} \;\mathbb{I}(\hat{y}(\mathbf{x}^t)\neq c),
\end{align}
where $\mathbb{I}(\cdot)$ is the indicator function.
Building upon the GMM formulated in Eq.~(\ref{eqn:gmm}), we now derive the optimal threshold as follow:
\begin{align}
    L &= \frac{1}{2} \sum_{c={k, u}} \,\sum_{y(\mathbf{x}^t)=c} p(\mathbf{x}^t|y(\mathbf{x}^t)=c) \;\mathbb{I}(\hat{y}(\mathbf{x}^t) \neq c) \notag\\
      &= \frac{1}{2} \sum_{c={k, u}} \,\sum_{y(\mathbf{x}^t)=c} p(m|y=c) \;\mathbb{I}(\hat{y}(\mathbf{x}^t) \neq c) \notag\\
      &= \frac{1}{2} \sum_{c={k, u}} \,\sum_{y(\mathbf{x}^t)=c} \phi(m|\hat{\mu}_c, \hat{\sigma}_c) \;\mathbb{I}(\hat{y}(\mathbf{x}^t) \neq c). \notag
\end{align}
Let assume that we employ a threshold $\gamma$ for unknown detection:
\begin{align}
\label{eqn:pred}
\hat{y}(\mathbf{x}^t) &= \left\{
\begin{aligned}
 &k, &\text{if$\quad m=\text{UniMS}(\mathbf{x}^t) \ge \gamma$}\\
 &u, &\text{if$\quad m=\text{UniMS}(\mathbf{x}^t) < \gamma$}
\end{aligned}
\right. .
\end{align}
Then, we have:
\begin{equation}
\begin{aligned}
\min_{\gamma} \quad L = \frac{1}{2} [ \,&\sum_{y(\mathbf{x}^t)=k, m < \gamma} \phi(m|\hat{\mu}_k, \hat{\sigma}_k)  \\ + 
&\sum_{y(\mathbf{x}^t)=u, m \ge \gamma} \phi(m|\hat{\mu}_u, \hat{\sigma}_u) ]. 
\end{aligned}
\end{equation}
When interpreted geometrically, we should aim to minimize the overlapping area between the two components of the GMM under the condition of uniform mixture weights.
Consequently, we need only assign uniform mixture weights and then use their intersection point as the threshold $\gamma$.

\begin{figure}[!t]
	\centering
	\includegraphics[width=0.8\columnwidth]{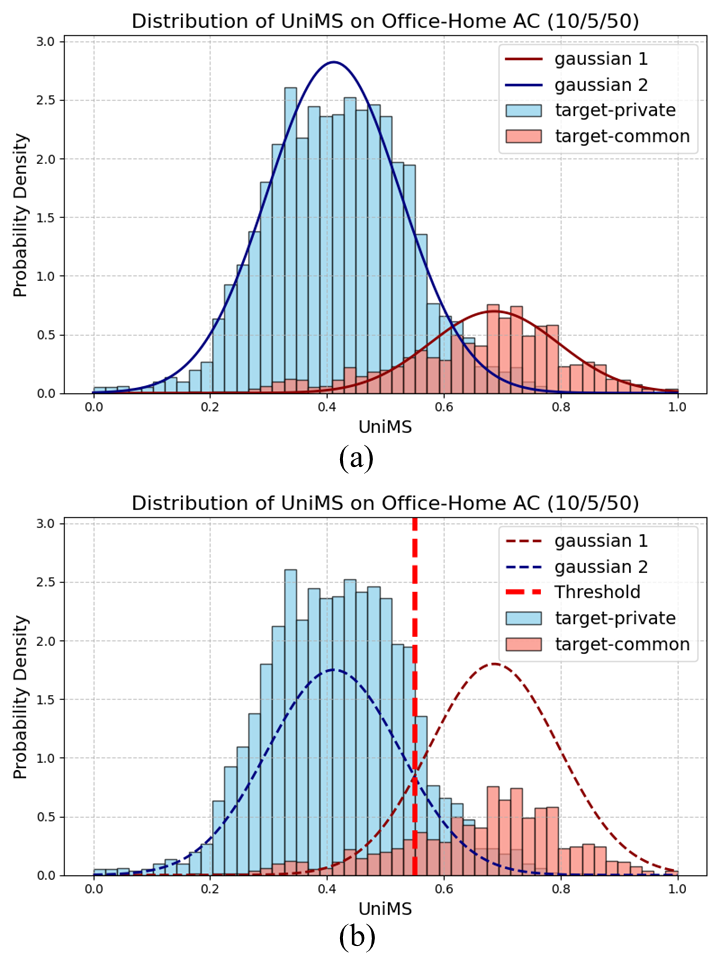}
	\caption{\small (a) Probability density distribution of UniMS and the estimated GMM by Eq.~(\ref{eqn:gmm_ours}), using data from AC in Office-Home (OPDA, i.e., 10/5/50). (b) Adjusted GMM featuring equal mixture weights, alongside the derived threshold (depicted as a red dotted line).}
	\label{fig:gmm}
\end{figure}

\subsubsection{Visualizations.}
To visually demonstrate the impact of our design on threshold selection, we conducted experiments on the Office-Home dataset.
As illustrated in Figure~\ref{fig:gmm} (a), our GMM effectively models the probability density distributions of both known and unknown classes with UniMS as the scoring function.
Additionally, Figure~\ref{fig:gmm} (b) illustrates that the threshold derived from the GMM with equal mixture weights strikes a good balance between known and unknown classes. This threshold retains a significant portion of target-common samples while minimally sacrificing the accuracy in identifying unknown classes.

\begin{algorithm}[!h]
\caption{TASC Algorithm}
\label{alg:TASC}
\DontPrintSemicolon
\KwIn{Image encoder $f$, text encoder $g$ and their parameters $\theta$. Source class names $\mathcal{T}^s$ and nouns $\mathcal{T}^{nouns}$. Unlabeled target data $\mathcal{D}^t$}
\KwOut{Nouns $\mathcal{T}$, hidden state vector $\mathbf{r}$, adapted model parameters $\theta$.}
\KwInit{Initialize $\mathcal{T}$ according to \mbox{Eq.~(8)}, set $\mathbf{r}$ to an all-ones vector.}
// \textit{Stage 1: Greedy Search-bsed Discrete Optimization}
    \For{$m=1$ {\bfseries to} $N_{outer}$}
    {
      \For{$i=1$ {\bfseries to} $K_0$}
        {
        \eIf{$i \leq |\mathcal{C}_s|$} 
          {
            Using only $r_i$ as the candidate. \;
            Obtain target domain prototypes $\mu$ using the current predictions based on $\mathcal{T}^\mathbf{r}$. \;
            \If{$(h(g(t_i^s);\boldsymbol{\mu},\tau)) < \gamma_{ent}$}  
            {$r_i\leftarrow 1$, continue}
          }
          {
          Randomly select $n_c$ candidates $\mathcal{T}^{c}$ from $\mathcal{T}^{nouns}$, including $t_i$. \;
          }
        Find the optimal $t^{\prime*}$ from the $n_c$ candidates and obtain $\mathcal{L}^{min}$ via Eq. (9).\;
        Calculate the loss $\mathcal{L}^{dis}$ when $r_i=0$. \;
// \textit{Update the discrete variables} \;
        $\mathcal{T} \leftarrow \mathcal{T}_{i|t^{\prime *}}.$ \;
        \eIf{$\mathcal{L}^{min}<\mathcal{L}^{dis}$} 
          {
          $\mathbf{r} \leftarrow \mathbf{r}_{i|1}$.
          }
          {
          $\mathbf{r} \leftarrow \mathbf{r}_{i|0}$.
          }
        }
    }
Obtain and fix the optimal $\mathcal{T}^*$ and $\mathbf{r}^*$. \;
// \textit{Stage 2: Model Refinement} \;
\For{$epoch=1$ {\bfseries to} $N_{epoch}$}
{
\For{mini-batch \textbf{in} $N_{epoch}$}
{
Calculate the loss in Eq.~(12). \;
// \textit{Update the continuous variables} \;
Update $\theta$ using Gradient Descent. \;
}
}
Obtain the optimal $\theta^*$. \;

\KwRet{$\mathcal{T}^*, \mathbf{r}^*, \theta^*$}
\end{algorithm}


\section{TASC Algorithm}
To rigorously describe our two-stage solution algorithm, we provide Algorithm~\ref{alg:TASC}. Note that all notations are consistent with the main text.

\begin{figure}[!t]
	\centering
	\includegraphics[width=1.0\columnwidth]{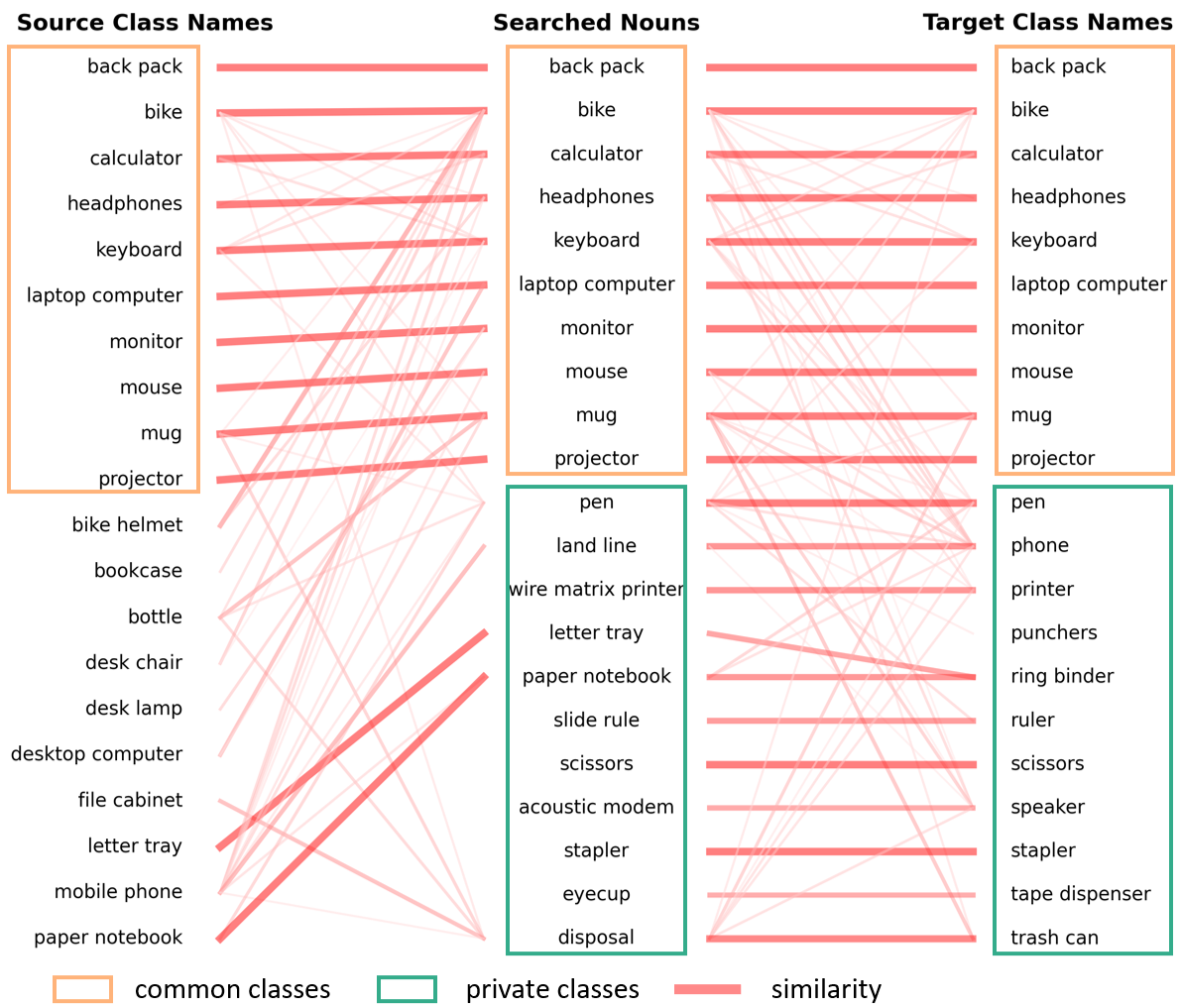}
	\caption{\small Presented from left to right are the source domain class names $\mathcal{T}^s$, the searched nouns $\mathcal{T}^{*\mathbf{r}^{*}}$, and the ground truth target class names $\mathcal{T}^t$. The similarity between each text pair is conveyed through the transparency and thickness of the connecting lines between them. The results are obtained from WD in Office dataset under OPDA (10/10/11) scenario. }
	\label{fig:nouns}
\end{figure}

Additionally, we have selected an example to visually showcase the results of our greedy search-based discrete optimization, as shown in Figure~\ref{fig:nouns}.

\section{Discussion on Recent Work}
\subsection{Universal Domain Adaptation.}
As we mentioned in the main paper, we propose leveraging the text representation space to facilitate developing simple and robust UniDA algorithms, which has not been fully explored.
One may question that a few existing works have explored using CLIP as the backbone for solving the UniDA problem, such as~\cite{unida_uniood,unida_COCA}.
Here, we emphasize that these methods~\cite{unida_uniood,unida_COCA} simply replace previous non-VLMs backbones with CLIP~\cite{CLIP}, neglecting to fully exploit the cross-modality matching capabilities.
\cite{unida_uniood} explored the foundation models for UniDA and proposed CLIP distillation, which focuses on distilling target knowledge from the CLIP models and calibration.
\cite{unida_COCA} aims to utilize the CLIP to reduce the labeling cost on source samples in SF-UniDA.
Besides,~\cite{unida_COCA} introduce the semantic one-vs-all clustering module, which utilizes the text embeddings as the classifier's weights, for detecting unknown samples.
Both papers overlook the fact that CLIP provides decent representations for open-world samples, and accurately representing the semantics of target-private classes with text embeddings is achievable.
Regarding our work, as we discussed in the main paper, we summarize and address the issues with existing clustering-based UniDA methods. 
We propose that uniformly representing the image semantics of both domains in the text representation space can facilitate the development of simple and robust UniDA algorithms.
This is the core idea of our method, which significantly differs from existing works~\cite{unida_uniood,unida_COCA}.

Recently,~\cite{unida_lead} address the SF-UniDA without tedious tuning thresholds or time-consuming iterative clustering, which decouples features into two components to identify target-private data.
For comparison, it should be noted that our approach also does not rely on the hand-crafted threshold hyper-parameter as discussed above. 
On the other hand,~\cite{unida_lead} still addresses the DA problem under category shift in the unconstrained continuous image representation space, while our method leveraging the semantically meaningful and discrete text representation space.

\subsection{Category Discovery and Image Clustering.}
The problem tackled by our method shares similarities with NCD~\cite{han2019learning} and GCD~\cite{vaze2022generalized} in certain respects, such as discovering the categories of the unlabeled dataset and estimating the number of categories. 
However, it is crucial to clarify that the key challenge in UniDA lies in the detection and alignment of common classes. 
Our framework not only discovers private classes in the target domain but also suppresses private classes from the source domain and retains common classes, thereby addressing the category shift problem.

Besides, existing works such as~\cite{unida_UniOT,jahan2024unknown} also perform category discovery.
However, these methods are unable to handle broader class shift scenarios. 
Additionally, they fail to fully exploit the semantics of open classes after category discovery, for achieving the better separability of known and unknown samples.
In contrast, we first explicitly model the category shift and then embed it into the similarity score for more robust unknown detection, (\ie, UniMS).
Notably, the second item in UniMS, which utilizes the semantics of target-private classes, \ie~MS-t w/ $\mathbf{ent}^t$, provides complementary information, resulting in higher performance.

On the other hand,~\cite{SCD} proposed a new task of obtaining class names for unlabeled datasets.
Compared to it, our method considers more practical application scenarios, \ie, transferring knowledge from a source domain to unlabeled target domain  under both a wide range of category shift and domain shift.
Besides, some methods~\cite{cai2023semantic,li2023image} also resort to the external knowledge from text modality.
The main differences between~\cite{cai2023semantic,li2023image} and us lie in the following aspects: we searched the optimal set of discrete semantic centers from text modality representation space, while they use semantic centers composed of multiple text embeddings; we address domain shift and class shift issues, while they tackle the problem of image clustering.
Notably, the former enables us to fine-tune the text encoder for better addressing the gap between modalities, since we represent the semantics via the class names.
Moreover, we ensure an appropriate semantic granularity by leveraging a semantically meaningful and discrete space.

\begin{table}[t]
\centering
\setlength{\tabcolsep}{5pt}
\resizebox{0.98\columnwidth}{!}{
\begin{tabular}{l|cccc}
\toprule
\multirow{2}{*}{Dataset} & \multicolumn{4}{c}{Class Split($\lvert\mathcal{C}\rvert/ \lvert\bar{\mathcal{C}}_s\rvert/ \lvert\bar{\mathcal{C}}_t\rvert$)}  \\
\cmidrule{2-5} & OPDA  & ODA  & PDA  & CDA\\ 
\midrule
Office & 10/10/11 & 10/0/11 & 10/21/0 & 31/0/0\\
Office-Home & 10/5/50 & 25/0/40 & 25/40/0 & 65/0/0\\
VisDA & 6/3/3 & -  & - & - \\
DomainNet & 150/50/145   & -       & -  & -\\ 
\bottomrule
\end{tabular}
}
\caption{Details of class split. $\mathcal{C}$, $\bar{\mathcal{C}}_s$, and $\bar{\mathcal{C}}_t$ denotes the common class, the source-private class, and the target-private class, respectively.}
\label{tab:class_split}
\end{table}

\section{More Results}
\label{sec:more_results}

\subsection{Detailed Results.}
Following existing works~\cite{unida_GLC,unida_UAN}, we evaluate our method on four different category shift scenarios to demonstrate its robustness, and the detailed class split in these scenarios are summarized in Table A1.
Due to space limitations, we did not provide detailed results of various methods on Office and Office-Home under 4 different scenarios in the main paper.
Here, we list all results for careful comparison, as shown in Tables A2 to A9.

\subsection{More Evaluation Metrics.}
To provide a more detailed demonstration of our method's performance, we report multiple metrics here.
In OPDA and ODA scenarios, we first calculate the per-class accuracy $a_{\mathcal{C}}$ on common classes, the per-class accuracy ``$a_{\mathcal{C}}$ w/o unk'' on common classes without the unknown detection mechanism in Eq.~(\ref{eqn:pred}) and the accuracy $a_{\bar{\mathcal{C}}_t}$ on a single unknown class.
Besides, following~\cite{unida_UniOT}, we obtain the NMI~\cite{mcdaid2011normalized} to measure the quality of target-private clustering via performing K-means on target-private samples.
Base on these metrics, the H-score~\cite{unida_CMU} and H$^3$-score~\cite{unida_UniOT} can be introduced as:
\begin{gather}
    \text{H-score}=\frac{2}
    {1/a_{\mathcal{C}} + 1/a_{\bar{\mathcal{C}}_t}},\\
    \text{H$^3$-score}=\frac{3}
    {1/a_{\mathcal{C}} + 1/a_{\bar{\mathcal{C}}_t} + 1/\text{NMI}}.
\end{gather}
Moreover, the \mbox{AUROC} of UniMS is utilized in order to evaluate the performance of target-private samples detection.
In PDA and CDA scenarios, we report the classification accuracy over all target samples following~\cite{unida_DCC,unida_GLC}.
The results of our method with these metrics are shown in~\cref{table:office_OPDA,table:office_ODA,table:office_PDA,table:office_CDA,table:officehome_OPDA,table:officehome_ODA,table:officehome_PDA,table:officehome_CDA,table:domainnet_OPDA,table:visda_OPDA}.

\begin{table}[htpb]
  \begin{center}
  \setlength{\tabcolsep}{3pt}
  \resizebox{\columnwidth}{!}{
  \begin{tabular}{l|ccccccc}
  \toprule
    &
    \multicolumn{1}{c}{AD} &
    \multicolumn{1}{c}{AW} &
    \multicolumn{1}{c}{DA} &
    \multicolumn{1}{c}{DW} &
    \multicolumn{1}{c}{WA} &
    \multicolumn{1}{c}{WD} &
    \multicolumn{1}{c}{Avg}
    \\  
\midrule
SO$^{\star}$ & 72.21 & 71.33 & 80.10 & 80.34 & 72.41 & 78.16 & 75.76 \\
DANCE$^{\star}$ & 95.08 & 82.49 & 87.07 & 94.62 & 83.79 & 94.89 & 89.65 \\
OVANet$^{\star}$ & 90.29 & 85.36 & \textbf{92.80} & \textbf{95.63} & \textbf{93.82} & 94.75 & \textbf{92.11} \\
UniOT$^{\star}$ & 90.10 & 84.38 & 90.78 & 94.24 & 90.77 & 90.31 & 90.10 \\
Ours & \textbf{96.37} & \textbf{86.48} & 88.44 & 89.27 & 90.99 & \textbf{96.45} & 91.33 \\
    \bottomrule
    \end{tabular}
  }
\caption{Comparison between our method and SOTAs on Office in OPDA scenarios.}
  \label{table:all_office_OPDA}
  \end{center}
  \end{table}

\begin{table}[htpb]
  \begin{center}
  \setlength{\tabcolsep}{3pt}
  \resizebox{\columnwidth}{!}{
  \begin{tabular}{l|ccccccc}
  \toprule
    &
    \multicolumn{1}{c}{AD} &
    \multicolumn{1}{c}{AW} &
    \multicolumn{1}{c}{DA} &
    \multicolumn{1}{c}{DW} &
    \multicolumn{1}{c}{WA} &
    \multicolumn{1}{c}{WD} &
    \multicolumn{1}{c}{Avg}
    \\  
\midrule
SO$^{\star}$ & 86.09 & 87.96 & 81.60 & 88.02 & 78.25 & 78.99 & 83.49 \\
DANCE$^{\star}$ & 95.83 & \textbf{94.76} & 79.40 & 98.02 & 76.56 & \textbf{99.42} & 90.67 \\
OVANet$^{\star}$ & 93.12 & 92.27 & 94.15 & 98.29 & 91.44 & 97.77 & 94.51 \\
UniOT$^{\star}$ & 91.96 & 87.61 & 92.65 & \textbf{98.62} & 93.38 & 98.09 & 93.72 \\
Ours & \textbf{98.77} & 94.18 & \textbf{94.80} & 95.34 & \textbf{94.85} & 98.37 & \textbf{96.05} \\
    \bottomrule
    \end{tabular}
  }
  \caption{Comparison between our method and SOTAs on Office in ODA scenarios.}
  \label{table:all_office_ODA}
  \end{center}
  \end{table}

\begin{table}[htpb]
  \begin{center}
  \setlength{\tabcolsep}{3pt}
  \resizebox{\columnwidth}{!}{
  \begin{tabular}{l|ccccccc}
  \toprule
    &
    \multicolumn{1}{c}{AD} &
    \multicolumn{1}{c}{AW} &
    \multicolumn{1}{c}{DA} &
    \multicolumn{1}{c}{DW} &
    \multicolumn{1}{c}{WA} &
    \multicolumn{1}{c}{WD} &
    \multicolumn{1}{c}{Avg}
    \\  
\midrule
SO$^{\star}$ & 92.00 & 91.30 & 94.07 & \textbf{99.67} & 93.76 & \textbf{100.00} & 95.13 \\
DANCE$^{\star}$ & 87.89 & 87.47 & 80.74 & \textbf{99.67} & 73.01 & \textbf{100.00} & 88.13 \\
OVANet$^{\star}$ & 92.89 & 91.47 & 94.67 & 98.64 & 93.55 & \textbf{100.00} & 95.20 \\
UniOT$^{\star}$ & 73.28 & 69.98 & 71.41 & 74.88 & 66.65 & 70.87 & 71.18 \\
Ours & \textbf{93.63} & \textbf{94.92} & \textbf{95.82} & 99.66 & \textbf{96.14} & 97.45 & \textbf{96.27} \\
    \bottomrule
    \end{tabular}
  }  \caption{Comparison between our method and SOTAs on Office in PDA scenarios.}
  \label{table:all_office_PDA}
  \end{center}
  \end{table}

\begin{table}[htpb]
  \begin{center}
  \setlength{\tabcolsep}{3pt}
  \resizebox{\columnwidth}{!}{
  \begin{tabular}{l|ccccccc}
  \toprule
    &
    \multicolumn{1}{c}{AD} &
    \multicolumn{1}{c}{AW} &
    \multicolumn{1}{c}{DA} &
    \multicolumn{1}{c}{DW} &
    \multicolumn{1}{c}{WA} &
    \multicolumn{1}{c}{WD} &
    \multicolumn{1}{c}{Avg}
    \\  
\midrule
SO$^{\star}$ & 88.81 & 86.95 & 75.28 & 98.86 & 74.03 & 99.71 & 87.27 \\
DANCE$^{\star}$ & 89.45 & 89.99 & 75.25 & \textbf{99.02} & 71.36 & \textbf{100.00} & 87.51 \\
OVANet$^{\star}$ & 88.95 & 85.32 & 73.59 & 98.60 & 74.04 & 99.85 & 86.73 \\
UniOT$^{\star}$ & 88.33 & \textbf{90.85} & 77.42 & 98.54 & 78.55 & 99.85 & 88.93 \\
Ours & \textbf{90.96} & 88.18 & \textbf{82.75} & 97.23 & \textbf{82.07} & 99.20 & \textbf{90.07} \\
    \bottomrule
    \end{tabular}
  }  \caption{Comparison between our method and SOTAs on Office in CDA scenarios.}
  \label{table:all_office_CDA}
  \end{center}
  \end{table}

\clearpage
\begin{table*}[htpb]
  \begin{center}

  \setlength{\tabcolsep}{5pt}
  \resizebox{\textwidth}{!}{
  \begin{tabular}{l|ccccccccccccc}
  \toprule
    &
    \multicolumn{1}{c}{AC} &
    \multicolumn{1}{c}{AP} &
    \multicolumn{1}{c}{AR} &
    \multicolumn{1}{c}{CA} &
    \multicolumn{1}{c}{CP} &
    \multicolumn{1}{c}{CR} &
    \multicolumn{1}{c}{PA} &
    \multicolumn{1}{c}{PC} &
    \multicolumn{1}{c}{PR} &
    \multicolumn{1}{c}{RA} &
    \multicolumn{1}{c}{RC} &
    \multicolumn{1}{c}{RP} &
    \multicolumn{1}{c}{Avg}
    \\  
\midrule
SO$^{\star}$ & 67.44 & 70.87 & 74.04 & 83.52 & 78.30 & 81.15 & 82.99 & 76.56 & 79.58 & 78.52 & 72.70 & 68.50 & 76.18 \\
DANCE$^{\star}$ & 78.18 & 81.66 & 82.23 & 86.03 & 89.42 & 81.48 & \textbf{89.41} & 83.16 & 86.58 & 87.27 & 81.08 & 79.75 & 83.85 \\
OVANet$^{\star}$ & 79.05 & 84.75 & 85.69 & 85.76 & \textbf{89.93} & 88.24 & 86.14 & 80.06 & 90.09 & 85.58 & 80.39 & 84.64 & 85.03 \\
UniOT$^{\star}$ & 83.08 & 87.72 & 90.76 & 83.34 & 86.72 & 89.72 & 81.91 & 83.44 & 91.15 & 82.76 & 83.61 & 91.31 & 86.29 \\
Ours & \textbf{86.26} & \textbf{89.69} & \textbf{91.09} & \textbf{89.79} & 88.50 & \textbf{93.11} & 88.65 & \textbf{85.50} & \textbf{91.91} & \textbf{90.12} & \textbf{86.79} & \textbf{91.90} & \textbf{89.44} \\
    \bottomrule
    \end{tabular}
  }
    \caption{Comparison between our method and SOTAs on Office-Home in OPDA scenarios.}
  \label{table:all_officehome_OPDA}
  \end{center}
  \end{table*}

\begin{table*}[htpb]
  \begin{center}

  \setlength{\tabcolsep}{5pt}
  \resizebox{\textwidth}{!}{
  \begin{tabular}{l|ccccccccccccc}
  \toprule
    &
    \multicolumn{1}{c}{AC} &
    \multicolumn{1}{c}{AP} &
    \multicolumn{1}{c}{AR} &
    \multicolumn{1}{c}{CA} &
    \multicolumn{1}{c}{CP} &
    \multicolumn{1}{c}{CR} &
    \multicolumn{1}{c}{PA} &
    \multicolumn{1}{c}{PC} &
    \multicolumn{1}{c}{PR} &
    \multicolumn{1}{c}{RA} &
    \multicolumn{1}{c}{RC} &
    \multicolumn{1}{c}{RP} &
    \multicolumn{1}{c}{Avg}
    \\  
\midrule
SO$^{\star}$ & 59.81 & 60.59 & 63.92 & 71.35 & 62.88 & 64.01 & 75.03 & 69.71 & 71.24 & 65.81 & 55.33 & 55.61 & 64.61 \\
DANCE$^{\star}$ & 64.48 & 71.78 & 66.55 & 75.79 & 47.83 & 53.17 & 78.17 & 72.17 & 70.98 & 72.58 & 56.86 & 54.45 & 65.40 \\
OVANet$^{\star}$ & 73.00 & 76.05 & 77.10 & 75.97 & 80.07 & 76.87 & 75.71 & 72.85 & 81.65 & 77.41 & 71.85 & 75.44 & 76.16 \\
UniOT$^{\star}$ & \textbf{77.84} & 83.31 & 86.29 & 79.15 & 83.14 & 84.14 & 71.82 & 72.29 & 84.73 & 77.13 & 76.04 & 87.28 & 80.26 \\
Ours & 76.15 & \textbf{88.76} & \textbf{87.75} & \textbf{82.40} & \textbf{87.89} & \textbf{86.13} & \textbf{81.22} & \textbf{77.12} & \textbf{88.47} & \textbf{81.83} & \textbf{78.46} & \textbf{88.78} & \textbf{83.75} \\
    \bottomrule
    \end{tabular}
  }
    \caption{Comparison between our method and SOTAs on Office-Home in ODA scenarios.}
  \label{table:all_officehome_ODA}
  \end{center}
  \end{table*}

\begin{table*}[htpb]
  \begin{center}

  \setlength{\tabcolsep}{5pt}
  \resizebox{\textwidth}{!}{
  \begin{tabular}{l|ccccccccccccc}
  \toprule
    &
    \multicolumn{1}{c}{AC} &
    \multicolumn{1}{c}{AP} &
    \multicolumn{1}{c}{AR} &
    \multicolumn{1}{c}{CA} &
    \multicolumn{1}{c}{CP} &
    \multicolumn{1}{c}{CR} &
    \multicolumn{1}{c}{PA} &
    \multicolumn{1}{c}{PC} &
    \multicolumn{1}{c}{PR} &
    \multicolumn{1}{c}{RA} &
    \multicolumn{1}{c}{RC} &
    \multicolumn{1}{c}{RP} &
    \multicolumn{1}{c}{Avg}
    \\  
\midrule
SO$^{\star}$ & 75.85 & 84.58 & 88.79 & 76.32 & 82.52 & 83.64 & 78.06 & 75.25 & 86.94 & 84.03 & 76.13 & 89.01 & 81.76 \\
DANCE$^{\star}$ & \textbf{81.00} & 88.59 & 91.26 & 78.73 & 83.00 & 86.82 & 75.16 & \textbf{82.95} & 90.88 & 87.55 & 76.22 & \textbf{93.06} & 84.60 \\
OVANet$^{\star}$ & 74.75 & 84.49 & 88.18 & 78.37 & 84.77 & 83.48 & 80.17 & 75.37 & 86.87 & 83.16 & 74.68 & 88.77 & 81.92 \\
UniOT$^{\star}$ & 69.60 & 76.67 & 80.65 & 69.72 & 77.33 & 80.35 & 69.90 & 66.41 & 80.88 & 76.54 & 69.00 & 80.88 & 74.83 \\
Ours & 79.70 & \textbf{93.11} & \textbf{94.04} & \textbf{90.17} & \textbf{92.77} & \textbf{93.15} & \textbf{89.26} & 82.33 & \textbf{93.71} & \textbf{90.54} & \textbf{78.81} & 92.83 & \textbf{89.20} \\
    \bottomrule
    \end{tabular}
  }
    \caption{Comparison between our method and SOTAs on Office-Home in PDA scenarios.}
  \label{table:all_officehome_PDA}
  \end{center}
  \end{table*}

\begin{table*}[htpb]
  \begin{center}

  \setlength{\tabcolsep}{5pt}
  \resizebox{\textwidth}{!}{
  \begin{tabular}{l|ccccccccccccc}
  \toprule
    &
    \multicolumn{1}{c}{AC} &
    \multicolumn{1}{c}{AP} &
    \multicolumn{1}{c}{AR} &
    \multicolumn{1}{c}{CA} &
    \multicolumn{1}{c}{CP} &
    \multicolumn{1}{c}{CR} &
    \multicolumn{1}{c}{PA} &
    \multicolumn{1}{c}{PC} &
    \multicolumn{1}{c}{PR} &
    \multicolumn{1}{c}{RA} &
    \multicolumn{1}{c}{RC} &
    \multicolumn{1}{c}{RP} &
    \multicolumn{1}{c}{Avg}
    \\  
\midrule
SO$^{\star}$ & 70.48 & 81.75 & 86.98 & 77.10 & 84.70 & 85.38 & 74.26 & 68.03 & 85.95 & 81.85 & 70.38 & 90.87 & 79.81 \\
DANCE$^{\star}$ & 72.91 & 83.08 & 88.28 & 79.32 & 85.73 & 86.75 & 77.78 & 70.97 & 87.44 & 83.32 & 73.20 & 92.05 & 81.74 \\
OVANet$^{\star}$ & 69.54 & 81.31 & 85.64 & 78.36 & 85.64 & 85.45 & 75.74 & 68.40 & 87.13 & 82.32 & 69.50 & 91.04 & 80.01 \\
UniOT$^{\star}$ & 72.59 & 85.53 & 87.78 & 79.93 & 87.32 & 86.27 & 77.27 & 71.68 & 87.81 & 82.56 & 73.81 & 92.02 & 82.05 \\
Ours & \textbf{74.62} & \textbf{91.71} & \textbf{92.15} & \textbf{86.86} & \textbf{91.91} & \textbf{91.90} & \textbf{87.19} & \textbf{75.65} & \textbf{92.04} & \textbf{88.05} & \textbf{75.95} & \textbf{92.88} & \textbf{86.74} \\
    \bottomrule
    \end{tabular}
  }
    \caption{Comparison between our method and SOTAs on Office-Home in CDA scenarios.}
  \label{table:all_officehome_CDA}
  \end{center}
  \end{table*}

\clearpage
\begin{table*}[htpb]
  \begin{center}

  \setlength{\tabcolsep}{10pt}
  \resizebox{\textwidth}{!}{
  \begin{tabular}{l|ccccccc}
  \toprule
    &
    \multicolumn{1}{c}{AD} &
    \multicolumn{1}{c}{AW} &
    \multicolumn{1}{c}{DA} &
    \multicolumn{1}{c}{DW} &
    \multicolumn{1}{c}{WA} &
    \multicolumn{1}{c}{WD} &
    \multicolumn{1}{c}{Avg}
    \\  
\midrule
$a_{\mathcal{C}}$ w/o unk & 99.11 & 98.27 & 96.94 & 99.10 & 96.84 & 100.00 & 98.38 \\
$a_{\mathcal{C}}$ & 97.33 & 94.21 & 93.33 & 97.74 & 94.07 & 100.00 & 96.11 \\
$a_{\bar{\mathcal{C}}_t}$ & 95.43 & 79.93 & 84.04 & 82.16 & 88.11 & 93.14 & 87.13 \\
NMI & 88.98 & 86.67 & 66.30 & 86.03 & 67.40 & 87.06 & 80.40 \\
AUROC & 99.34 & 95.02 & 92.62 & 97.69 & 96.32 & 98.41 & 96.57 \\
H-score & 96.37 & 86.48 & 88.44 & 89.27 & 90.99 & 96.45 & 91.33 \\
H$^3$-score & 93.77 & 86.54 & 79.58 & 88.16 & 81.48 & 93.10 & 87.11 \\
    \bottomrule
    \end{tabular}
  }
    \caption{Results (\%) of our method on Office in OPDA scenarios.}
  \label{table:office_OPDA}
  \end{center}
  \end{table*}

\begin{table*}[htpb]
  \begin{center}

  \setlength{\tabcolsep}{10pt}
  \resizebox{\textwidth}{!}{
  \begin{tabular}{l|ccccccc}
  \toprule
    &
    \multicolumn{1}{c}{AD} &
    \multicolumn{1}{c}{AW} &
    \multicolumn{1}{c}{DA} &
    \multicolumn{1}{c}{DW} &
    \multicolumn{1}{c}{WA} &
    \multicolumn{1}{c}{WD} &
    \multicolumn{1}{c}{Avg}
    \\  
\midrule
$a_{\mathcal{C}}$ w/o unk & 100.00 & 100.00 & 97.05 & 99.77 & 97.55 & 100.00 & 99.06 \\
$a_{\mathcal{C}}$ & 98.68 & 92.17 & 90.86 & 94.77 & 91.45 & 97.88 & 94.30 \\
$a_{\bar{\mathcal{C}}_t}$ & 98.86 & 96.28 & 99.11 & 95.91 & 98.51 & 98.86 & 97.92 \\
NMI & 89.66 & 87.88 & 65.80 & 84.93 & 66.67 & 89.78 & 80.79 \\
AUROC & 99.92 & 98.54 & 98.86 & 98.37 & 99.00 & 99.92 & 99.10 \\
H-score & 98.77 & 94.18 & 94.80 & 95.34 & 94.85 & 98.37 & 96.05 \\
H$^3$-score & 95.53 & 91.98 & 82.66 & 91.60 & 83.14 & 95.33 & 90.04 \\
    \bottomrule
    \end{tabular}
  }
    \caption{Results (\%) of our method on Office in ODA scenarios.}
  \label{table:office_ODA}
  \end{center}
  \end{table*}

\begin{table*}[htpb]
  \begin{center}

  \setlength{\tabcolsep}{10pt}
  \resizebox{\textwidth}{!}{
  \begin{tabular}{l|ccccccc}
  \toprule
    &
    \multicolumn{1}{c}{AD} &
    \multicolumn{1}{c}{AW} &
    \multicolumn{1}{c}{DA} &
    \multicolumn{1}{c}{DW} &
    \multicolumn{1}{c}{WA} &
    \multicolumn{1}{c}{WD} &
    \multicolumn{1}{c}{Avg}
    \\  
\midrule
Acc & 93.63 & 94.92 & 95.82 & 99.66 & 96.14 & 97.45 & 96.27 \\
    \bottomrule
    \end{tabular}
  }
    \caption{Results (\%) of our method on Office in PDA scenarios.}
  \label{table:office_PDA}
  \end{center}
  \end{table*}

\begin{table*}[htpb]
  \begin{center}

  \setlength{\tabcolsep}{10pt}
  \resizebox{\textwidth}{!}{
  \begin{tabular}{l|ccccccc}
  \toprule
    &
    \multicolumn{1}{c}{AD} &
    \multicolumn{1}{c}{AW} &
    \multicolumn{1}{c}{DA} &
    \multicolumn{1}{c}{DW} &
    \multicolumn{1}{c}{WA} &
    \multicolumn{1}{c}{WD} &
    \multicolumn{1}{c}{Avg}
    \\  
\midrule
Acc & 90.96 & 88.18 & 82.75 & 97.23 & 82.07 & 99.20 & 90.07 \\
    \bottomrule
    \end{tabular}
  }
    \caption{Results (\%) of our method on Office in CDA scenarios.}
  \label{table:office_CDA}
  \end{center}
  \end{table*}

\clearpage
\begin{table*}[htpb]
  \begin{center}

  \setlength{\tabcolsep}{5pt}
  \resizebox{\textwidth}{!}{
  \begin{tabular}{l|ccccccccccccc}
  \toprule
    &
    \multicolumn{1}{c}{AC} &
    \multicolumn{1}{c}{AP} &
    \multicolumn{1}{c}{AR} &
    \multicolumn{1}{c}{CA} &
    \multicolumn{1}{c}{CP} &
    \multicolumn{1}{c}{CR} &
    \multicolumn{1}{c}{PA} &
    \multicolumn{1}{c}{PC} &
    \multicolumn{1}{c}{PR} &
    \multicolumn{1}{c}{RA} &
    \multicolumn{1}{c}{RC} &
    \multicolumn{1}{c}{RP} &
    \multicolumn{1}{c}{Avg}
    \\  
\midrule
$a_{\mathcal{C}}$ w/o unk & 90.99 & 98.99 & 99.26 & 96.67 & 98.79 & 99.02 & 96.24 & 91.37 & 99.02 & 96.36 & 91.47 & 98.99 & 96.43 \\
$a_{\mathcal{C}}$ & 81.76 & 94.51 & 96.87 & 86.85 & 98.07 & 95.57 & 86.17 & 80.35 & 91.88 & 87.69 & 82.94 & 97.05 & 89.97 \\
$a_{\bar{\mathcal{C}}_t}$ & 91.29 & 85.34 & 85.95 & 92.93 & 80.64 & 90.77 & 91.27 & 91.35 & 91.95 & 92.70 & 91.01 & 87.27 & 89.37 \\
NMI & 75.61 & 90.06 & 86.72 & 81.27 & 89.31 & 88.23 & 81.09 & 75.97 & 88.16 & 81.57 & 72.95 & 89.72 & 83.39 \\
AUROC & 93.20 & 97.61 & 97.02 & 96.86 & 98.22 & 98.08 & 95.35 & 93.38 & 96.64 & 97.08 & 94.42 & 98.51 & 96.36 \\
H-score & 86.26 & 89.69 & 91.09 & 89.79 & 88.50 & 93.11 & 88.65 & 85.50 & 91.91 & 90.12 & 86.79 & 91.90 & 89.44 \\
H$^3$-score & 82.39 & 89.82 & 89.58 & 86.76 & 88.77 & 91.43 & 85.98 & 82.07 & 90.63 & 87.08 & 81.63 & 91.16 & 87.27 \\
    \bottomrule
    \end{tabular}
  }
    \caption{Results (\%) of our method on Office-Home in OPDA scenarios.}
  \label{table:officehome_OPDA}
  \end{center}
  \end{table*}

\begin{table*}[htpb]
  \begin{center}

  \setlength{\tabcolsep}{5pt}
  \resizebox{\textwidth}{!}{
  \begin{tabular}{l|ccccccccccccc}
  \toprule
    &
    \multicolumn{1}{c}{AC} &
    \multicolumn{1}{c}{AP} &
    \multicolumn{1}{c}{AR} &
    \multicolumn{1}{c}{CA} &
    \multicolumn{1}{c}{CP} &
    \multicolumn{1}{c}{CR} &
    \multicolumn{1}{c}{PA} &
    \multicolumn{1}{c}{PC} &
    \multicolumn{1}{c}{PR} &
    \multicolumn{1}{c}{RA} &
    \multicolumn{1}{c}{RC} &
    \multicolumn{1}{c}{RP} &
    \multicolumn{1}{c}{Avg}
    \\  
\midrule
$a_{\mathcal{C}}$ w/o unk & 88.08 & 97.05 & 97.37 & 93.61 & 96.36 & 96.91 & 92.64 & 87.98 & 97.15 & 92.59 & 87.29 & 96.44 & 93.62 \\
$a_{\mathcal{C}}$ & 66.46 & 88.56 & 87.22 & 74.61 & 88.33 & 83.67 & 77.04 & 70.34 & 88.68 & 73.92 & 68.70 & 87.86 & 79.62 \\
$a_{\bar{\mathcal{C}}_t}$ & 89.14 & 88.96 & 88.30 & 92.00 & 87.45 & 88.73 & 85.87 & 85.35 & 88.26 & 91.63 & 91.45 & 89.71 & 88.91 \\
NMI & 73.54 & 90.05 & 87.70 & 81.66 & 90.32 & 89.26 & 82.55 & 73.88 & 88.63 & 83.98 & 72.66 & 89.71 & 83.66 \\
AUROC & 86.13 & 96.45 & 95.61 & 92.70 & 95.10 & 94.55 & 91.23 & 86.03 & 95.90 & 92.01 & 88.87 & 96.07 & 92.55 \\
H-score & 76.15 & 88.76 & 87.75 & 82.40 & 87.89 & 86.13 & 81.22 & 77.12 & 88.47 & 81.83 & 78.46 & 88.78 & 83.75 \\
H$^3$-score & 75.26 & 89.19 & 87.74 & 82.15 & 88.69 & 87.15 & 81.66 & 76.01 & 88.52 & 82.53 & 76.42 & 89.09 & 83.70 \\
    \bottomrule
    \end{tabular}
  }
    \caption{Results (\%) of our method on Office-Home in ODA scenarios.}
  \label{table:officehome_ODA}
  \end{center}
  \end{table*}

\begin{table*}[htpb]
  \begin{center}

  \setlength{\tabcolsep}{7pt}
  \resizebox{\textwidth}{!}{
  \begin{tabular}{l|ccccccccccccc}
  \toprule
    &
    \multicolumn{1}{c}{AC} &
    \multicolumn{1}{c}{AP} &
    \multicolumn{1}{c}{AR} &
    \multicolumn{1}{c}{CA} &
    \multicolumn{1}{c}{CP} &
    \multicolumn{1}{c}{CR} &
    \multicolumn{1}{c}{PA} &
    \multicolumn{1}{c}{PC} &
    \multicolumn{1}{c}{PR} &
    \multicolumn{1}{c}{RA} &
    \multicolumn{1}{c}{RC} &
    \multicolumn{1}{c}{RP} &
    \multicolumn{1}{c}{Avg}
    \\  
\midrule
Acc & 79.70 & 93.11 & 94.04 & 90.17 & 92.77 & 93.15 & 89.26 & 82.33 & 93.71 & 90.54 & 78.81 & 92.83 & 89.20 \\
    \bottomrule
    \end{tabular}
  }
    \caption{Results (\%) of our method on Office-Home in PDA scenarios.}
  \label{table:officehome_PDA}
  \end{center}
  \end{table*}

\begin{table*}[htpb]
  \begin{center}

  \setlength{\tabcolsep}{7pt}
  \resizebox{\textwidth}{!}{
  \begin{tabular}{l|ccccccccccccc}
  \toprule
    &
    \multicolumn{1}{c}{AC} &
    \multicolumn{1}{c}{AP} &
    \multicolumn{1}{c}{AR} &
    \multicolumn{1}{c}{CA} &
    \multicolumn{1}{c}{CP} &
    \multicolumn{1}{c}{CR} &
    \multicolumn{1}{c}{PA} &
    \multicolumn{1}{c}{PC} &
    \multicolumn{1}{c}{PR} &
    \multicolumn{1}{c}{RA} &
    \multicolumn{1}{c}{RC} &
    \multicolumn{1}{c}{RP} &
    \multicolumn{1}{c}{Avg}
    \\  
\midrule
Acc & 74.62 & 91.71 & 92.15 & 86.86 & 91.91 & 91.90 & 87.19 & 75.65 & 92.04 & 88.05 & 75.95 & 92.88 & 86.74 \\
    \bottomrule
    \end{tabular}
  }
    \caption{Results (\%) of our method on Office-Home in CDA scenarios.}
  \label{table:officehome_CDA}
  \end{center}
  \end{table*}

\clearpage
\begin{table*}[htpb]
  \begin{center}

  \setlength{\tabcolsep}{10pt}
  \resizebox{\textwidth}{!}{
  \begin{tabular}{l|ccccccc}
  \toprule
    &
    \multicolumn{1}{c}{PR} &
    \multicolumn{1}{c}{PS} &
    \multicolumn{1}{c}{RP} &
    \multicolumn{1}{c}{RS} &
    \multicolumn{1}{c}{SP} &
    \multicolumn{1}{c}{SR} &
    \multicolumn{1}{c}{Avg}
    \\  
\midrule
$a_{\mathcal{C}}$ w/o unk & 87.59 & 74.27 & 73.21 & 73.39 & 74.28 & 87.45 & 78.37 \\
$a_{\mathcal{C}}$ & 79.29 & 66.17 & 63.40 & 65.14 & 64.04 & 79.46 & 69.58 \\
$a_{\bar{\mathcal{C}}_t}$ & 81.79 & 73.61 & 75.97 & 74.07 & 76.16 & 83.61 & 77.53 \\
NMI & 85.47 & 66.71 & 73.14 & 66.27 & 74.75 & 85.25 & 75.27 \\
AUROC & 91.50 & 83.46 & 84.93 & 84.09 & 84.38 & 91.91 & 86.71 \\
H-score & 80.52 & 69.69 & 69.12 & 69.32 & 69.57 & 81.48 & 73.28 \\
H$^3$-score & 82.10 & 68.67 & 70.41 & 68.27 & 71.22 & 82.70 & 73.90 \\
    \bottomrule
    \end{tabular}
  }
    \caption{Results (\%) of our method on DomainNet in OPDA scenarios.}
  \label{table:domainnet_OPDA}
  \end{center}
  \end{table*}

\begin{table*}[htpb]
  \begin{center}

  \setlength{\tabcolsep}{10pt}

  \begin{tabular}{l|c}
  \toprule
    &
    \multicolumn{1}{c}{SR}
    \\  
\midrule
$a_{\mathcal{C}}$ w/o unk & 94.58 \\
$a_{\mathcal{C}}$ & 90.94 \\
$a_{\bar{\mathcal{C}}_t}$ & 89.79 \\
NMI & 89.46 \\
AUROC & 96.67 \\
H-score & 90.36 \\
H$^3$-score & 90.06 \\
    \bottomrule
    \end{tabular}
  \caption{Results (\%) of our method on VisDA in OPDA scenarios.}
  \label{table:visda_OPDA}
  \end{center}
  \end{table*}

\clearpage

\end{document}